\documentclass[journal,onecolumn,12pt, oneside,draft]{IEEEtran}
\usepackage{array}
\usepackage{longtable}
\newcolumntype{L}[1]{>{\raggedright\let\newline\\\arraybackslash\hspace{0pt}}m{#1}}

\usepackage[inline]{enumitem}

\usepackage[final]{graphicx}
\graphicspath{{.//}}
\usepackage{ragged2e}
\usepackage{makecell}
\usepackage{tikz}
\usetikzlibrary{backgrounds}
\usetikzlibrary{arrows}
\tikzstyle{arrow} = [thick,->,>=stealth]
\tikzset{
	block/.style = {draw, thick, minimum width=1.7cm, minimum height=1.2cm, node distance=3cm},
	down/.style={yshift=-7em},
	up/.style={yshift=7em},
	right/.style={xshift=15em},
	downright/.style={xshift=12em, yshift = -7em},
	downleft/.style={xshift=-12em, yshift = -7em}
}
\tikzset{
	circ/.style={circle, fill=black, inner sep=2pt, node contents={}}
}
\usetikzlibrary{shapes,arrows}
\tikzstyle{line}=[draw] 
\usepackage{epstopdf}
\usetikzlibrary{calc}
\usetikzlibrary{arrows.meta}
\usepackage{tikz-3dplot}
\usetikzlibrary{patterns}
\usetikzlibrary{tikzmark}
\usepackage{pgfplots}

\usepackage{calrsfs}
\DeclareMathAlphabet{\pazocal}{OMS}{zplm}{m}{n}

\usepackage{amssymb}
\usepackage{amsthm} 
\theoremstyle{definition}

\usepackage{amsmath}
\DeclareMathOperator*{\argmax}{arg\,max}

\usepackage{mathtools}

\DeclarePairedDelimiterX\Basics[1](){ #1}
\newtheorem{theorem}{Theorem}[section]

\newtheorem{lemma}[theorem]{Lemma}

\DeclareSymbolFont{matha}{OML}{txmi}{m}{it}
\DeclareMathSymbol{\varv}{\mathord}{matha}{118}
 \usepackage{relsize}
\usepackage{bm}
 \newcommand*\xor{\mathbin{\oplus}}
 
\usepackage{placeins}
\usepackage[noadjust]{cite}
\usepackage{hyperref}
\usepackage[caption=false,font=footnotesize]{subfig}
\title{Design of Capacity-Approaching Low-Density Parity-Check Codes using Recurrent Neural Networks}
\author{Eleni Nisioti and Nikolaos Thomos~\IEEEmembership{Senior~Member,~IEEE}
\thanks{E. Nisioti and N. Thomos are with the School of Computer Science and Electronic Engineering, University of Essex, Colchester, United Kingdom (e-mail:e.nisioti, nthomos@essex.ac.uk).} 
}

\begin{document}

\maketitle

\begin{abstract}
In this paper, we model Density Evolution (DE) using Recurrent Neural Networks (RNNs) with the aim of designing capacity-approaching Irregular Low-Density Parity-Check (LDPC) codes for binary erasure channels. In particular, we present a method for determining the coefficients of the degree distributions, characterizing the structure of an LDPC code. We refer to our RNN architecture as Neural Density Evolution (NDE) and determine the weights of the RNN that correspond to optimal designs by minimizing a loss function that enforces the properties of asymptotically optimal design, as well as the desired structural characteristics of the code. This renders the LDPC design process highly configurable, as constraints can be added to meet applications' requirements by means of modifying the loss function. In order to train the RNN, we generate data corresponding to the expected channel noise. We analyze the complexity and optimality of NDE theoretically, and compare it with traditional design methods that employ differential evolution. Simulations illustrate that NDE improves upon differential evolution both in terms of asymptotic performance and complexity. Although we focus on asymptotic settings, we evaluate designs found by NDE for finite codeword lengths and observe that performance remains satisfactory across a variety of channels.
\end{abstract}

\section{Introduction}\label{sec:intro}
The quality of communication over erroneous channels is degraded due to the presence of noise. It is known that, in order to ensure an arbitrarily small probability of error at the receiver, the rate of communication should not exceed the channel capacity, dictated by the level and nature of noise \cite{shannon}. Channel coding \cite{Luby:1998:ALD:276698.276756,2763,1638543} aims at adding redundancy to the communicated messages so that errors, introduced during transmission over a noisy channel, can be corrected at the receiver. Low-density parity-check (LDPC) codes \cite{1057683} are a prominent family of channel codes, the interest in which has been recently reignited due to the need of 5G for low complexity coding schemes that can recover messages at rates close to the channel capacity \cite{8316763}. In order to find well-performing channel codes for a variety of code-rates and message lengths, we are in need of efficient, effective and configurable design methods for this family of channel codes. 

\begin{figure}
\begin{minipage}[t]{0.45\textwidth}
	\centering
	\scalebox{0.7}{
	\begin{tikzpicture}
\node[draw, text width=2cm,align=center] (coding) at (0,0) {Coding};
\node[] (source) at ([xshift=-1.5cm,yshift=-0.5cm]coding) {$\bm{s}$};
\node[] (source_top) at ([xshift=-2.5cm]coding) {};
\node[draw, text width=2.5cm,align=center] (channel) at ([xshift=3cm]coding) {Transmission};
\node[] (code) at ([xshift=1.5cm,yshift=-0.5cm]coding) {$\bm{c}$};
\node[draw, text width=2cm,align=center] (decoding) at ([xshift=3cm]channel) {Decoding};
\node[draw, text width=2cm,align=center] (design) at ([yshift=-2cm]channel) {Code design};
\node[] (code_ch) at ([xshift=1.5cm,yshift=-0.5cm]channel) {$\hat{\bm{c}}$};
\node[] (source_ch) at ([xshift=1.5cm,yshift=-0.5cm]decoding) {$\tilde{\bm{s}}$};
\node[] (source_ch_top) at ([xshift=2.5cm]decoding) {};
\draw[->] (source_top) -- (coding) ;
\draw[->] (coding) -- (channel) ;
\draw[->] (channel) -- (decoding) ;
\draw[->] (decoding) -- (source_ch_top) ;
\draw[<->] (design) -| (decoding) ;
\draw[<->] (design) -| (coding) ;
\end{tikzpicture}}
	\caption{The communication pipeline: code design leverages information about the channel noise and code performance to find well-performing codes.}
	\label{fig:comm}
\end{minipage} \quad\quad\quad
\begin{minipage}[t]{0.45\textwidth}
	\centering
	\scalebox{0.7}{
	\begin{tikzpicture}
\node[draw, circle, text width=0.32cm,align=center] (vn1) at (0,0) {$v_1$};
\node[draw,circle,text width=0.32cm,align=center] (vn2) at ([xshift=1cm]vn1) {$v_2$};
\node[draw,circle,text width=0.32cm,align=center] (vn3) at ([xshift=1cm]vn2) {$v_3$};
\node[draw,circle,text width=0.32cm,align=center] (vn4) at ([xshift=1cm]vn3) {$v_4$};
\node[draw,circle,text width=0.32cm,align=center] (vn5) at ([xshift=1cm]vn4) {$v_5$};
\node[draw,circle,text width=0.32cm,align=center] (vn6) at ([xshift=1cm]vn5) {$v_6$};
\node[draw,text width=0.3cm,align=center] (cn1) at ([yshift=1.5cm, xshift=1.5cm]vn1) {$c_1$};
\node[draw,text width=0.3cm,align=center] (cn2) at ([xshift=1cm]cn1) {$c_2$};
\node[draw,text width=0.3cm,align=center] (cn3) at ([xshift=1cm]cn2) {$c_3$};
\draw (cn1) -- (vn1);
\draw (cn1) -- (vn2);
\draw (cn1) -- (vn3);
\draw (cn1) -- (vn4);
\draw (cn2) -- (vn3);
\draw (cn2) -- (vn4);
\draw (cn2) -- (vn6);
\draw (cn3) -- (vn1);
\draw (cn3) -- (vn4);
\draw (cn3) -- (vn5);
\end{tikzpicture}}
	\caption{The bipartite graph used to describe an LDPC code: the lower (upper) row consists of variable (check) nodes. An edge indicates that a codeword bit participates in a parity-check constraint.}
	\label{fig:bipartite}
\end{minipage}
\end{figure}

Without loss of generality, we assume messages generated by independent and identically distributed sources \cite{shannon}. Prior to transmission, source information bits are partitioned into non-overlapping message blocks of $k$ bits, and each block $\bm{s}$ is encoded by means of channel coding to form a codeword $\bm{c}$ of $n$ bits, with $n>=k$. The added bits aim at protecting the source message against errors introduced due to channel noise. Source messages can be recovered at the receiver's end by means of channel decoding, which maps the received codewords (error-contaminated) $\hat{\bm{c}}$ to an estimation of the source message $\tilde{\bm{s}}$ using, among other techniques, maximum likelihood estimation \cite{Richardson:2008:MCT:1795974}. Channel codes aim at recovering the original message at the decoder, i.e. $\tilde{\bm{s}}=\bm{s}$, with high probability, while at the same time adding as little redundancy as possible. This overall communication flow chart is presented in Fig. \ref{fig:comm}.

Low-density parity-check codes \cite{1057683} can be described in two ways: \begin{enumerate*}[label=(\roman*)]
    \item using the parity-check matrix $\bm{H}$, which determines the set of ``admissible'' codewords of a code, denoted as $\pazocal{C}(n,k)$. In particular, a codeword is admissible, if it belongs to the set of right null vectors of $\bm{H}$, i.e. $ \bm{x} \in \pazocal{C}(n,k) \text{ iff } \bm{H} \bm{c}^T=\bm{0}^{T}$. The parity-check matrix is related to the generator matrix $\bm{G}$, used to encode an information block $\bm{s}$ in order to generate a codeword $\bm{x}$ transmitted over the channel, with the formula: $\bm{G}\bm{H}^T=\bm{0}^T$;
    \item the Tanner graph, a bipartite graph of the form presented in Fig. \ref{fig:bipartite}, where each variable node is associated with a codeword bit and check nodes represent parity-check constraints. In the example presented in Fig. \ref{fig:bipartite}, codewords respect the following parity-check constraints:  $c_1 \xor c_2 \xor c_4 = 0 ,c_3 \xor c_4 \xor c_6 = 0, c_1 \xor _4 \xor c_5 =0$, where $\xor$ denotes the XOR function
\end{enumerate*} The relationship between the two representations is straightforward: variable nodes correspond to columns of $\bm{H}$ and check nodes to rows; an edge exists between two nodes if the corresponding entry in $\bm{H}$ is non-zero. The structure of LDPC codes is traditionally derived by sampling the degree distributions $\lambda(x)$ and $\rho(x)$ in order to determine the degrees\footnote{The degree of a node equals the number of edges emanating from it.} of the variable and check nodes respectively. As $\bm{H}$ is typically sparse, both encoding and decoding have low time complexity increasing linearly with the codeword length.


\begin{figure}
\begin{minipage}[t]{0.45\textwidth}
\begin{equation*}
     \resizebox{3cm}{1cm}{
$ \begin{bmatrix}
1 & 1 & 0 & 1 & 0 & 0\\
0 & 0 & 1 & 1 & 0 & 0 \\
1 & 0 & 0 & 1 & 1 & 0
\end{bmatrix}$
}
\end{equation*}
\caption{The parity-check matrix $\bm{H}$ corresponding to the bipartite graph of Fig. \ref{fig:bipartite}.}
    \label{fig:H}
\end{minipage}\quad\quad\quad
\begin{minipage}[t]{0.45\textwidth}
    \centering
    \scalebox{0.5}{
    \begin{tikzpicture}
	\node[draw,minimum width=2cm, minimum height=1.5cm] (search) at (0,0) {Search};
	\node[draw,minimum width=2cm, minimum height=1.5cm] (BP) at ([xshift=4cm]search) {\makecell{BP \\ or \\ DE}};
	\draw[arrow]  (search.north) -- ([yshift=1cm]search.north) -| node [xshift=-2cm, above] {\Large $\lambda(x), \rho(x)$}  (BP.north);
	\draw[arrow] (BP.south) -- ([yshift=-1cm]BP.south) -| node [xshift=2cm, below] {\Large $\epsilon^{\text{BP}}$}  (search.south);
\end{tikzpicture}}
    \caption{The traditional approach to designing irregular LDPC codes: the space of degree distributions is searched for candidate pairs $(\lambda(x),\rho(x))$, which are evaluated using Belief Propagation or Density Evolution.}
    \label{fig:design}
\end{minipage}
\end{figure}

Density Evolution (DE) is a recursive update formula that models the evolution of messages exchanged in a Belief propagation (BP) decoder, and is traditionally employed for the design of LDPC codes \cite{910578}. DE assumes that codewords have infinite length, which significantly simplifies the analysis, by converting the Tanner graph to a tree, where messages evolve independently from each other. In this work, we assume that codewords are transmitted over the binary erasure channel (BEC), introduced in \cite{1057464}, in alignment with the literature related to the analysis of DE \cite{910578,shokrollahi_7.7_nodate}. In the BEC, the probability of decoding failure at iteration $t+1$ under DE is given by $x_{t+1}(\epsilon)=\epsilon\lambda(1-(\rho(1-x_t)))$, where $x_t$ is the probability of failure at iteration $t$, $\rho(x)$ and $\lambda(x)$ are the degree distributions characterizing the considered code, and $\epsilon$ is the probability of erasure. The code design process is a two-step iterative procedure, illustrated in Fig. \ref{fig:design}. First, the space of degree distributions is explored to find degree distribution pairs for evaluation, a highly non-linear constraint satisfaction problem, traditionally solved using differential evolution \cite{910578}. Second, the selected pairs are evaluated using DE or BP by calculating $\epsilon^{\text{BP}}$, the maximum erasure probability that can be tolerated by the code.


By viewing code design as a non-linear optimization problem, we can alternatively leverage the framework of artificial neural networks (ANNs). Recurrent neural networks (RNNs) are ANNs employed when data exhibit time dependencies, and their dynamic behavior can be described by differential equations \cite{Pascanu:2013:DTR:3042817.3043083, 279181, 230622}. From this perspective, learning can be seen as an attempt to make the internal state of the network converge to the fixed point indicated by the labels provided in the training data \cite{Li:1994:LFP:185297.185299}. In this paper, we model the code design process as a supervised learning problem by mapping the recursive update equation of DE to an RNN architecture,  that we refer to as Neural Density Evolution (NDE). This RNN modeling constitutes an elegant way to incorporate $\lambda(x)$ and $\rho(x)$ into the design process simultaneously, as degree distribution coefficients are mapped to the weights of two consecutive layers of the neural network. The assumption of infinite codeword lengths, made by DE analysis, is also required by the proposed solution. Due to this assumption, we can also leverage the concentration theorem \cite{luby_practical_1997}, according to which the performance of the various realizations of the graph concentrates around their expected value. We, therefore, construct the parity-check matrix randomly by sampling the degree distributions and evaluate the designs based on ensemble averages. 

In order to find the weights of NDE that lead to capacity-approaching designs, we employ gradient-based optimizers \cite{DBLP:journals/corr/Ruder16}. Although gradient-based search was employed in \cite{910578} on the evaluation of $(\lambda(x),\rho(x))$ pairs using a BP decoder and heuristic search was coupled with DE, to the best of our knowledge, this is the first method that combines gradient-based optimization and DE analysis. Up to now, the best-performing designs in the literature have been found using differential evolution, a global optimization technique \cite{Ilonen2003}. However, our results prove that local optimization can lead to better designs, if appropriate measures are taken to avoid convergence to local optima of bad quality. This improvement over \cite{910578} can be attributed to the prohibitive complexity of global optimizers, which in practical settings (i.e. large codeword lengths and degree distribution degrees) fail to approach the theoretically optimal solution.  In Section \ref{sec:complexity}, we comment on the complexity of NDE and, in Section \ref{sec:efficiency}, we perform extensive simulations to compare the efficiency of NDE, in terms of time complexity, with that of code design using differential evolution.



Furthermore, we define a multi-objective loss function for NDE which reflects the requirements of optimal code design. In particular, we design loss terms to ensure that: \begin{enumerate*}[label=(\roman*)]
    \item the learned weights correspond to coefficients of valid degree distributions,
    \item the learned weights lead to the desired code-rate and density of the parity-check matrix $\bm{H}$, and, 
    \item the probability of decoding failure is zero
\end{enumerate*}. Our approach belongs to the realm of soft constraint optimization, while previous approaches that ensured valid degree distributions by projecting solutions on constraint-induced hyperplanes \cite{910578} employ hard constraints. Although soft constraints have been found to require special attention when balancing loss terms, they are far more computationally attractive than imposing hard constraints on the weights of the network \cite{DBLP:journals/corr/Marquez-NeilaSF17}. Simultaneously, our definition of the loss function ensures high configurability of NDE, the behavior of which can be further customized by introducing additional loss terms.

In order to train and evaluate NDE, we generate artificial data based on the target channel model and capacity. We, then, investigate the optimality of our solution, where we analyze DE as a dynamical system and comment on the ability of NDE to find capacity-achieving and capacity-approaching designs. To this aim, we leverage known results on the optimality of irregular LDPC codes, as well as properties of RNNs and gradient-descent optimizers \cite{910578}.

We experimentally confirm that NDE can be used to design capacity approaching codes for a variety of configurations, i.e. various code-rates and maximum degrees of $\lambda(x)$ and $\rho(x)$. We also examine the properties of the learned DE and draw meaningful conclusions on the effect that the number of decoding iterations has on the quality of the design. As part of our experimental analysis, we employ two types of graphs for describing the dynamic behavior of NDE: \begin{enumerate*}[label=(\roman*)]
    \item bifurcation diagrams, commonly used to visualize the behavior of dynamical systems \cite{230622,Pascanu:2013:DTR:3042817.3043083};
    \item a type of plot referred to as the \textit{graphical determination of threshold} in \cite{Richardson:2008:MCT:1795974}
\end{enumerate*}. Although during optimization we assume infinite codeword lengths and transmission over the BEC, we evaluate our designed codes on the transmission of moderate codeword lengths over the channel with additive white Gaussian noise (AWGN) noise. Our simulations confirm that the generated codes remain well-performing, which is in accordance to what was observed in \cite{910578}.  

In the rest of the paper, we first present related works in Section \ref{sec:related}. Relevant theoretical background is provided in Section \ref{sec:background}. In Section \ref{sec:NDE}, we present our main contribution, NDE, on the complexity of which we comment in Section \ref{sec:complexity}. We, then, proceed with an analysis of NDE's optimality in Section \ref{sec:optimality}. Simulations for evaluating NDE in asymptotic and finite settings are presented in Section \ref{sec:simulations}, and in Section \ref{sec:discussion} we conclude our work and comment on future directions.

\section{Related work}\label{sec:related}

Irregular LDPC codes \cite{luby_practical_1997} quickly acquired a prominent position in the family of channel codes due to their low complexity and their ability to transmit at rates close to capacity with low probability of error. In \cite{910578}, DE was leveraged for the design of irregular LDPC codes. First, optimal values for the coefficients of $\lambda(x)$ were found using a simple local search algorithm, where only a few non-zero terms were proven adequate for achieving good performance.  In addition, a derivative was formulated for hill-climbing algorithms and globally optimizing a linear interpolation of the loss function. The calculation of the latter required simulating the BP decoder and evaluating the bit-error-rate (BER) of the code during the design process. This differs from our approach, which, although also calculates derivatives as part of the back-propagation algorithm employed by the RNN, does not require information related to the simulated performance. Considering such information would be computationally expensive and would only ensure that the designed code will be optimal for a predefined code-length. The best-performing optimization technique in \cite{910578} was differential evolution, a global evolutionary optimizer often used for multivariate loss functions.  The work in \cite{910578} offered important insights into the nature of code design as an optimization problem, as it was observed that searching over the parameter space often gets trapped at critical points that are almost fixed points. Despite being capable of finding the optimal solution, differential evolution was also studied in \cite{Ilonen2003} as an optimizer for training ANNs and proved to be inferior, both in complexity and achieved performance.

In order to improve the performance of irregular LDPC codes under an asymptotic analysis, DE was extensively studied as a dynamical system in \cite{Richardson2004FIXEDPA, 910578}. These works contain a condition under which zero, the point representing a zero probability of decoding failure and, thus, corresponding to successful communication, is a stable fixed point of DE. It was also observed that zero is always a fixed point, but, for a specific degree distribution pair $(\lambda(x), \rho(x))$, additional fixed points can come into existence. In \cite{Richardson:2008:MCT:1795974}, it was also observed that a capacity-achieving code has an infinite number of fixed points when operating at the Shannon capacity, as BP decoding comes to a halt for any initial condition. Above the Shannon capacity, the fixed point at zero is unstable, so the probability of decoding error converges to a high value, which lies close to the initial point of evaluation of DE \cite{Richardson2004FIXEDPA}.

The dynamic properties of RNNs early on attracted the interest of the research community, as their dynamic behavior significantly differs from feedforward neural networks. In \cite{279181}, the long-standing observation that RNNs cannot learn long-term dependencies was attributed to the fact that these required that the gradient, whose calculation is necessary for back-propagation, vanishes or explodes. In \cite{230622}, a study of an RNN with a single neuron using the concept of bifurcation diagrams is presented.




The application of machine learning techniques in the area of channel codes has recently attracted significant interest. In contrast to our work, which concerns itself with the design of codes, the majority of recent contributions are concerned with decoding. In \cite{8054694}, an end-to-end communication system was viewed as an autoencoder, thus allowing for the encoding and decoding process to be jointly optimized. In \cite{8683463}, deep learning was employed to perform joint source and channel coding, with raw signal values being directly mapped to complex-valued transmitted signals, and results confirmed the advantages of this design for image communication and its resilience to noise mismatch. In \cite{DBLP:journals/corr/GruberCHB17}, channel decoding was viewed as a supervised classifier, employed to decode randomly constructed codes and polar codes, with the former failing to generalize to unseen codewords. This approach suffers from the curse of dimensionality, as the size of the data increases to prohibitive levels, even for deep learning solutions. Different from these works, which do not explicitly optimize the channel code but view it as an end-to-end system, we aim at leveraging properties of the code design process already established in the coding theory, such as the independence of the evolution of messages in a tree-structured Tanner graph and the concentration theorem. In \cite{8242643}, BP was mapped to a neural network and both a feedforward and a recurrent architecture was examined. The authors observed that the ability of correcting errors was improved, when compared to that of traditional BP, an observation attributed to the fact that learning led to finding values for the weights which obliterate short cycles, known for increasing the probability of BP failure. The neural network implementation of BP in \cite{8242643} resembles to some extend ours, as both schemes implement a message-passing update mechanism. In particular, DE is equivalent to BP under the assumption of infinite codeword lengths. However, we should note that our work is concerned code design, while the ANN in \cite{8242643} was used for decoding. In \cite{DBLP:journals/corr/abs-1901-05719}, the problem of error-correction code design was solved using both a reinforcement learning and a genetic programming solver and well-performing designs were found. However, these techniques required feedback from the decoder, which can be problematic if we consider the introduced delay, the increased time complexity of the design process and the probability of having erroneous feedback. Also, both solvers are not expected to adapt quickly to changing channel conditions, as they require extensive interaction to converge.



The classical definition of irregular LDPC codes has been extended in a variety of directions. In particular, the idea of viewing a code as a combination of smaller is considered essential for ensuring low complexity and versatility in 5G New Radio \cite{8316763}. We would like to note that our solution regards the code design problem and, hence, can be coupled with alternative definitions of LDPC codes. For example, NDE can be useful in improving the macroscopic statistical properties of LDPC codes in 5G New Radio, as these are predicted by DE, while its architecture can be extended to consider cascades of codes \cite{luby_practical_1997}.




\section{Background}\label{sec:background}
In this section, we provide some additional background on linear block codes and the design of irregular LDPC codes using DE.

\subsection{Linear block codes}
Let us assume an end-to-end communication system as the one illustrated in Fig. \ref{fig:comm}. Furthermore, let us consider that the source messages are binary vectors $\mathbf{s}=[s_1,...s_k]$, with $ s_i \in \{0,1\}$ and that all mathematical operations take place in $\mathbb{F}_2$. A channel code $C(n,k)$ is characterized by its code-rate $R = k/n$, which corresponds to the percentage of information bits $k$ in a codeword of length $n$. Channel decoding can be performed by means of maximum likelihood estimation, where an estimate $\tilde{\bm{c}}$ of the original codeword $\bm{c}$ is calculated as:%
\begin{align}\label{eq:map}
\tilde{\bm{c}} = \argmax_{\bm{c} \in \pazocal{C}(n,k)} p_{\hat{\bm{c}}|\bm{c}}(\hat{\bm{c}}|\bm{c})     
\end{align}
Recall that $\pazocal{C}(n,k)$ is a set containing all admissible codewords and $p_{\hat{\bm{c}}|\bm{c}}(\hat{\bm{c}}|\bm{c}) $ represents the conditional probability of observing the codeword $\hat{\bm{c}}$ (after transmission through the channel) given that the codeword $\bm{c}$ was transmitted. (\ref{eq:map}) is valid under the assumption that all codewords are equiprobable and is traditionally solved using BP.




\subsection{Designing irregular LDPC codes}
The structure of an irregular LDPC code, as presented in Figs. \ref{fig:bipartite} and \ref{fig:H}, can be derived from the degree distributions, which have the following form:%
\begin{equation}
	\lambda(x) = \sum_{i=0}^{\lambda_{\text{max}}} \lambda_i x^{i+1} \qquad \rho(x)= \sum_{i=0}^{\rho_{\text{max}}} \rho_i x^{i+1}
	\label{eq:distr}
\end{equation}
where $\lambda_{\text{max}}(\rho_{\text{max}})$ is the maximum allowed number of edges emanating from a variable (check) node. An edge in the graph has a lower (upper) degree $i$ (see Fig. \ref{fig:bipartite}), if it is connected to a lower (upper) node of degree $i$. The objective of code design is to determine the variables $\lambda_i$ ($\rho_i$). Polynomial distributions of the above form are valid degree distributions if they have non-negative coefficients and their sum is equal to 1. In addition, the degree distributions need to respect the rate constraints of the code, i.e. $n \cdot \bar{\lambda}=k \cdot \bar{\rho}$, where $\bar{\lambda}$ and $\bar{\rho}$ denote the average degree of the distributions.

Once the degree distributions have been found, LDPC codes can be constructed by randomly sampling them. Through $\lambda(x)$ and $\rho(x)$, a family of codes is determined, as there are multiple $\bm{H}$ matrices that meet the constraints. However, this does not affect the quality of the design under the assumption of infinite codeword lengths, due to the concentration theorem \cite{luby_practical_1997}. A trivial way to build the bipartite graph from $\lambda(x)$ and $\rho(x)$ is to sample these distributions to determine the number of edges emanating from a variable (check) node, and then, randomly connect the edges. We also remove 4-cycles from $\bm{H}$  \cite{nocycles}, which significantly improves the achieved BER, as has been reported in the related literature \cite{910577}.

\subsection{Density Evolution}
As we briefly discussed in Section \ref{sec:intro}, DE has an iterative form, given in (\ref{eq:DE}) for BEC. Let $\epsilon \in [0,1]$ denote the probability of a bit being erased during transmission due to channel noise. The probability of decoding failure at the next decoding iteration under DE is given by:%
\begin{equation} \label{eq:DE}
x_{t+1}(\epsilon) = \epsilon \lambda(1-\rho(1-x_t)),
\end{equation}
where $x_t$ denotes the probability of failure at iteration $t$. DE can be used to assess the design indicated by a ($\lambda(x), \rho(x)$) pair without requiring to evaluate the performance of the channel code in a simulated channel. The evaluation begins by considering an initial point $x_0$, which is typically set to $\epsilon$, as this corresponds to the probability of failure prior to the beginning of decoding. Due to the assumption of infinite codeword lengths and i.i.d. source message bits, we only examine the evolution of the decoding for one variable node, and assume that the same occurs independently for the other nodes.

\begin{figure}
	\begin{minipage}[t]{0.45\textwidth}
				\scalebox{0.22}{
    \includegraphics{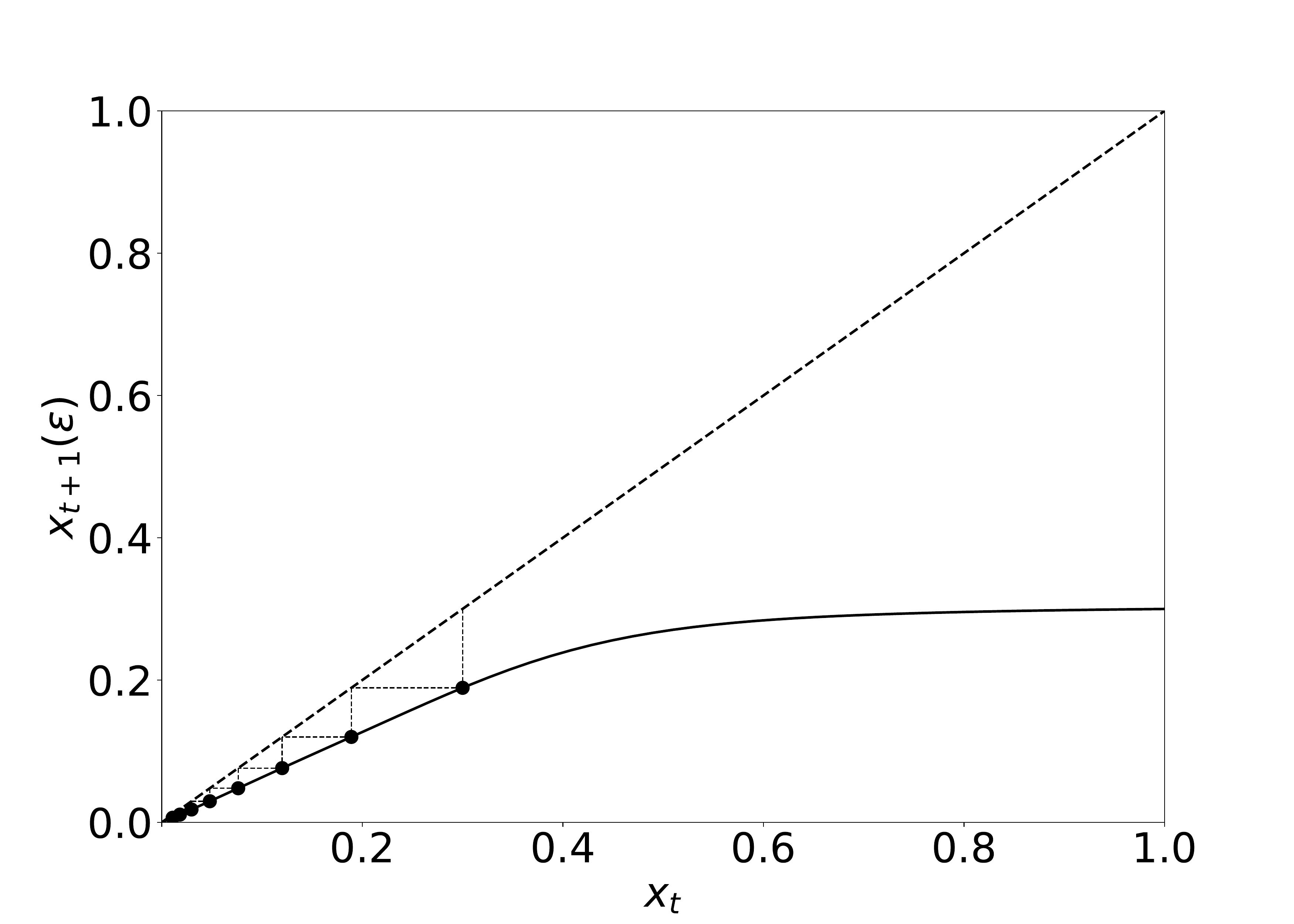}}
    		\caption{The evolution of the probability of decoding failure $x_t$ using Density Evolution for code-rate $R=0.5$, $\epsilon=0.3$ and degree distributions with average degrees $\bar{\lambda}=3.5662 $ and $\bar{\rho} = 7.1856$.}
		\label{fig:DE}
	\end{minipage} \quad\quad\quad
	\begin{minipage}[t]{0.45\textwidth}
	\centering
		\begin{tikzpicture}
\node[draw] (win) at (0,0) {$W_{\text{in}}$};
\node[draw] (wrec) at ([xshift=2cm]win) {$W_{\text{rec}}$};
\node[draw] (bias) at ([xshift=-2cm]win) {$b$};
\node[] (input) at ([xshift=-0.7cm, yshift=-0.5cm]win) {$u_t$};
\node[] (state) at ([xshift=-0.7cm, yshift=-0.5cm]wrec) {$x_t$};
\node[] (loss) at ([xshift=0.7cm, yshift=-0.5cm]wrec) {$L_t$};
\node[] (placeholder) at ([xshift=1cm]wrec) {};
\draw[->] (bias) -- (win);
\draw[->] (win) -- (wrec);
\draw[->] (wrec) -- (placeholder);
\path (wrec) edge [loop above] (wrec);
\end{tikzpicture}
		\caption{Schematic representation of a recurrent neural network, where the input $u_t$ and bias $b$ is applied once and, then, the state $x_t$ evolves recursively for a pre-defined number of iterations, until the final output is used to calculate the loss $L_t$.}
		\label{fig:RNN}
	\end{minipage}
\end{figure}

In Fig. \ref{fig:DE}, we illustrate how the probability of failure, $x_t$, evolves with decoding iterations for a specific ($\lambda(x), \rho(x)$) pair using this formula. We observe that $x_t$ approaches 0 within a small number of iterations, as the probability of erasure $(\epsilon=0.3)$ was lower than the code-rate ($R=0.5$).

\section{Neural Density Evolution} \label{sec:NDE}
In this section, we describe Neural Density Evolution, our proposed solution for LDPC code design. First, we define the update formula of an RNN that mimics the dynamics of DE. Then, we map this formula to our RNN architecture and describe the data generation process. Finally, we define a loss function that guarantees the desired code characteristics and ensures capacity-approaching performance.

\subsection{Mapping DE to an RNN}

We can view an RNN as a dynamical system, described by the following difference equation:%
\begin{align}\label{eq:RNN}
x_t = W_{\text{rec}} \sigma(x_{t-1}) + W_{\text{in}}u_t + b
\end{align}
with $W_{\text{rec}}$ denoting the recurrent weight matrix, $W_{\text{in}}$ the input weight matrix, $b$ a constant bias term and $\sigma(\cdot)$ representing the sigmoid activation function. An RNN accepts an input $u_t$ and is characterized by its state $x_t$, updated at each learning iteration. We illustrate this behavior of an RNN in Fig. \ref{fig:RNN}.

As our intention is that of modeling DE, we need to adjust the general RNN update equation, as expressed in (\ref{eq:RNN}), so that it agrees with the DE formula presented in (\ref{eq:DE}). For this reason, we drop the terms related to the bias, replace the sigmoid activation function $\sigma(\cdot)$ with the identity function and substitute the input weight matrix with the identity matrix. In addition, we replace the addition between the first two terms in (\ref{eq:RNN}) with a multiplication. Thus, the difference equation of our RNN is:%
\begin{align}\label{eq:RNN2}
x_t = W_{\text{rec}} (x_{t-1}) u_t
\end{align}

These transformations do not affect the ability of the RNN to find good solutions, as ANNs have been found well-performing for a variety of activation functions and architectures \cite{Goodfellow:2016:DL:3086952}. As is customary in the machine learning community, we refer to the elements in $ W_{\text{rec}}$ as the parameters of the learning algorithm, while we use the term hyperparameters to describe the parameters of the learning algorithm that are set prior to training, such as the learning rate, number of epochs and length of the unfolded RNN.

\subsection{Defining the RNN architecture}\label{sec:architecture}
\begin{figure}
	\centering
	\scalebox{0.8}{
	\begin{tikzpicture}[every node/.style={minimum size = 0.3cm,inner sep=0pt}]
\node[draw,circle] (input) at (0,0) {};
\node[draw,circle,fill=gray!30] (trans1) at ([xshift=2cm, yshift=1cm]input) {};
\node[draw,circle,fill=gray!30] (trans2) at ([xshift=2cm]input) {};
\node[draw,circle,fill=gray!30] (trans3) at ([xshift=2cm, yshift=-1cm]input) {};
\node[draw,circle] (factor1) at ([xshift=2cm]trans2) {};
\node[draw,circle,fill=gray!30] (trans4) at ([xshift=2cm, yshift=1cm]factor1) {};
\node[draw,circle,fill=gray!30] (trans5) at ([xshift=2cm]factor1) {};
\node[draw,circle,fill=gray!30] (trans6) at ([xshift=2cm, yshift=-1cm]factor1) {};
\node[draw,circle,fill=gray!30] (trans7) at ([xshift=2cm, yshift=2.5cm]factor1) {};
\node[draw,circle,fill=gray!30] (trans8) at ([xshift=2cm, yshift=-2.5cm]factor1) {};
\node[draw,circle] (var1) at ([xshift=2cm]trans5) {};
\node[minimum size=0cm] (phantom1) at ([yshift=-3cm]var1) {};
\node[minimum size=0cm] (phantom2) at ([yshift=-3cm]input) {};
\draw[->] (input.east) -- node [near end,above, xshift=-0.5cm ] {\tiny $(1-x)^0$} (trans1);
\draw[->] (input.east) -- node [near end,above ] {\tiny $(1-x)^1$}  (trans2);
\draw[->] (input.east) -- node [near end,below,xshift=-0.5cm  ] {\tiny $(1-x)^2$}  (trans3);
\draw[->] (trans1) -- node [near start,above,yshift=0.1cm ] {\tiny $\rho_1$}  (factor1);
\draw[->] (trans2) -- node [near start,above] {\tiny $\rho_2$}  (factor1);
\draw[->] (trans3) -- node [near start,below,yshift=-0.1cm ] {\tiny $\rho_3$}  (factor1);
\draw[->] (factor1.east) -- node [near end,above, yshift=0.2cm] {\tiny $(1-x)^1$} (trans4);
\draw[->] (factor1.east) -- node [near end,above,] {\tiny $(1-x)^2$} (trans5);
\draw[->] (factor1.east) -- node [near end,above, xshift=0.25cm,yshift=0.1cm ] {\tiny $(1-x)^3$} (trans6);
\draw[->] (factor1.east) -- node [near end,above, xshift=-0.5cm, yshift=0.1cm ] {\tiny $(1-x)^0$} (trans7);
\draw[->] (factor1.east) -- node [near end,below,xshift=-0.5cm, yshift=-0.2cm] {\tiny $(1-x)^4$} (trans8);
\draw[->] (trans4) -- node [near start,above,yshift=0.1cm ] {\tiny $\lambda_2$} (var1);
\draw[->] (trans5) -- node [near start,above ] {\tiny $\lambda_3$} (var1);
\draw[->] (trans6) -- node [near start,below,yshift=-0.1cm ] {\tiny $\lambda_4$} (var1);
\draw[->] (trans7) -- node [near start,above,yshift=0.2cm] {\tiny $\lambda_1$} (var1);
\draw[->] (trans8) -- node [near start,below,yshift=-0.2cm ] {\tiny $\lambda_5$} (var1);
\draw[->] (var1.south) -| (phantom1) -- (phantom2) |- (input.south);
\draw [dashed] (4,3.5) -- (4,-3.5);
\node[] (layer1) at (2,3.5) {Layer 1};
\node[] (layer2) at (6,3.5) {Layer 2};
\end{tikzpicture}}
	\caption{NDE architecture for an irregular LDPC code with $\lambda_{\text{max}}=5$ and $\rho_{\text{max}}=3$. Each layer employs a set of weights, which form a vector multiplied with the transformed state of the RNN. }
	\label{fig:network}
\end{figure}
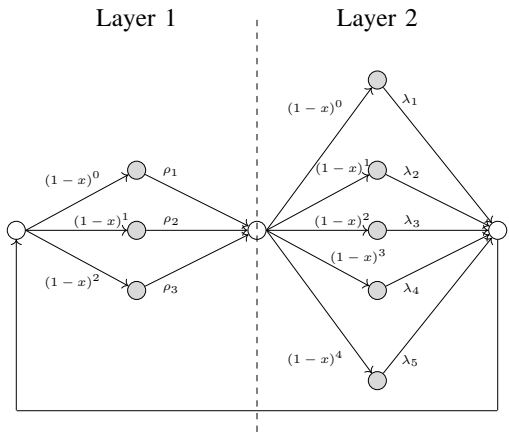

The trainable weights of NDE correspond to the coefficients of the degree distributions, defined in (\ref{eq:distr}). When compared to feed-forward architectures, an RNN bears the advantage that weights are shared among corresponding layers, which means that the learning algorithm needs to optimize a small number of parameters. The proposed RNN architecture is presented in Fig. \ref{fig:network}, where the coefficients of $\rho(x)$ have been mapped to the weights of the first layer of the RNN and the coefficients of $\lambda(x)$ to the weights of the second layer. We restrict the length of the unfolded RNN to a few layers, as this helps maintain low training complexity. As the length of the RNN coincides with the number of decoding iterations performed by BP, we expect that the performance of the design will not be affected, as, when BP converges, this usually occurs early in the decoding process \cite{Kok:2006:CMR:1248547.1248612}.%


\subsection{Generating training data} \label{sec:generate}
In order to train the RNN in a supervised manner, we need labeled data, of the form $(\text{features},\text{class})$. As we aim at designing channel codes for BEC, we present how we generate data for this type of channel. We define as features the inputs of DE, namely $\epsilon$, which represents the probability of a bit erasure and $x$, the point of evaluation of DE. The class of our training data indicates the desired output of DE, which is 0, i.e., the fixed point that corresponds to the probability that the data has been recovered after decoding with no errors for $ \epsilon< \epsilon^{\text{Sh}}$, where $\epsilon^{\text{Sh}}$ denotes the channel capacity, and a positive constant in the range $(\epsilon^{\text{Sh}},1)$ otherwise. As we observe in the bifurcation diagram presented in  Fig. \ref{fig:bifurcNDE}, this constant is very close to $\epsilon^{\text{Sh}}$. Note that, the analysis employed by DE is independent of the transmitted codeword \cite{Luby:1998:ARP:314613.314722}, we therefore assume transmission of the all-zero codeword.

In order to select an appropriate range of values for $\epsilon$, we can consult the following Lemma \cite{Richardson:2008:MCT:1795974}:%
\begin{lemma}[Monotonicity with respect to channel \cite{Richardson:2008:MCT:1795974}] \label{lemma:monoton}
	Let $(\lambda(x), \rho(x))$ be a degree distribution pair and $\epsilon \in [0,1]$. If $x_t(\epsilon) \xrightarrow{t \rightarrow \infty} 0$, then $x_t(\epsilon^{\prime}) \xrightarrow{t \rightarrow \infty}  0$ for all $0 \leq \epsilon^{\prime} \leq \epsilon$.
	\label{lemma:monochannel}
\end{lemma}

It is, however, known that the performance of irregular LDPC codes is bounded away from capacity by a constant that depends on $\bar{\rho}$. In particular, the analysis in \cite{1077796} calculated this constant as $\delta_{\text{min}} = ( R^{\bar{\rho}-1} (1-R))/ (1 + R^{\bar{\rho}-1}(1-R))$. We denote as $\epsilon^{\text{BP}}_{\text{max}}$ the noise value that corresponds to $\delta_{\text{min}}$, which represents the maximum achievable noise value below which the probability of error using a BP decoder approaches 0, and can be calculated as $(1-\delta_{\text{min}}-R)/(1- \delta_{\text{min}})$. Thus, for our problem, $\epsilon^{\text{BP}}_{\text{max}}$, and not $\epsilon^{\text{Sh}}$, is considered the optimal solution.

Based on the preceding discussion, we can train the network to minimize BER for error probabilities close to $\epsilon^{\text{BP}}_{\text{max}}$ and expect BER to be zero for lower error probabilities. However, it is intuitive, and was confirmed by simulations, that creating data for lower erasure values can facilitate training. This idea was motivated by curriculum learning, a machine learning technique according to which an optimizer is presented with data of increasing difficulty in order to gradually manage to solve harder problems \cite{Bengio:2009:CL:1553374.1553380}.

Another important hyperparameter is the number of training examples, the choice of which also affects the number of training epochs. If not enough training examples are generated, then the erasure probabilities might not be representative of the channel and, if they are too many, training times will be long without any further improvements from the perspective of the code design. Also, the smaller the targeted Shannon capacity, the less training data we need, as with smaller erasure values there is less variation of codewords.

\subsection{Designing the loss function}\label{sec:lossfunction}
We pursue capacity-approaching designs by minimizing the mean squared error between the prediction coming from NDE and the class of the training data.  As is common in the literature, we consider that decoding is successful if the probability of error falls below a threshold, $o_{\text{low}}$. We therefore set the prediction $y$ to 0, if it is lower than $o_{\text{low}}$. Since the class of the training examples is zero for examples below capacity, the network will attempt to set all $\lambda_i$ and $\rho_i$ weights to 0. However, in order to describe valid degree distributions, the weights need to satisfy the following constraints:%
\begin{equation}\label{eq:constraints}
\begin{split}
	\sum_{i=0}^{\lambda_{\text{max}}} \lambda_i =1 \quad & \text{and} \quad 0 \leq \lambda_i \leq 1, \  \forall i \in [0, \cdots, \lambda_{\text{max}}-1]  \\
\sum_{i=0}^{\rho_{\text{max}}} \rho_i =1	\quad & \text{and} \quad 0 \leq \rho_i \leq 1, \  \forall i \in [0, \cdots, \rho_{\text{max}}-1] \\
	\end{split}	
\end{equation}
To enforce the above constraints we can use weight clipping, a technique commonly employed to avoid saturation of the signals traversing the RNN. However, if we separately clip each weight by value, then we change the direction of the gradient calculated by gradient descent during training, which will most probably harm the quality of back-propagation  \cite{Goodfellow-et-al-2016}. Furthermore, applying weight clipping on the norm of the weight vector would not lead to coefficients that lie in the $[0,1]$ range. We therefore enforce the constraints in (\ref{eq:constraints}) indirectly, by employing additional loss terms that penalize weight vectors whose sum is not one and their individual elements are not in the $[0,1]$ range:%
\begin{equation} \label{eq:normalize}
\begin{split}
   \Omega_{\lambda} &= \sum_{i=0}^{\lambda_{\text{max}}}{\min{(\lambda_i,0)}} + \sum_{i=0}^{\lambda_{\text{max}}}{\max{(\lambda_i,1)}} + |1-\sum_{i=0}^{\lambda_{\text{max}}}{\lambda_i}| \\
   \Omega_{\rho} &= \sum_{i=0}^{\rho_{\text{max}}}{\min{(\rho_i,0)}} + \sum_{i=0}^{\rho_{\text{max}}} {\max{(\rho_i,1)}} + |1-\sum_{i=0}^{\rho_{\text{max}}}{\rho_i}|
\end{split}
\end{equation} 

In order to ensure that our design satisfies the rate constraint, i.e. $\bar{\lambda}/\bar{\rho}=R$, we calculate the desired $\bar{\lambda}$ and $\bar{\rho}$, based on the target codeword size and density $d$ of $\bm{H}$, and employ the following additional loss terms: 
\begin{align} \label{eq:density}
	\Omega_{\bar{\lambda}} &= \bar{\lambda}_{\text{desired}} - \sum_{i=2}^{\lambda_{\text{max}}}{(i-1) \lambda_i} \\
	\Omega_{\bar{\rho}} &= \bar{\rho}_{\text{desired}} - \sum_{i=2}^{\rho_{\text{max}}}{(i-1) \rho_i} 
\end{align}
where $\bar{\lambda}_{\text{desired}}$ is calculated as $d \cdot k$ and $\bar{\rho}_{\text{desired}}$ as $d \cdot n$. The above terms facilitate the customization of NDE for varying structural requirements by setting $\bar{\lambda}_{\text{desired}}$ and $\bar{\lambda}_{\text{desired}}$ to their desired values.

To conclude, we formulate the total loss as calculated over all training examples $D_{\text{train}}$:%
\begin{equation}\label{eq:totalloss}
\begin{split}
L({y},{\hat{y}}) &= c_{\text{MSE}}\frac{1}{n}\sum_{i=0}^{D_{\text{train}}} ({y_i}-{\hat{y_i}})^2 + c_{\lambda} \Omega_{\lambda} \\ &+ c_{\rho} \Omega_{\rho} +  c_{\bar{\lambda}}\Omega_{\bar{\lambda}} + c_{\bar{\rho}}\Omega_{\bar{\rho}}
\end{split}
\end{equation}
where ${y_i}$ is the class, ${\hat{y_i}}$ is the prediction of the recurrent network and $c_{\text{MSE}}, c_{\lambda}, c_{\rho}, c_{\bar{\lambda}}$ and $c_{\bar{\rho}}$ are constant parameters that control the gravity of the different loss terms, the values of which are tuned during simulations. In Section \ref{sec:optimality}, we introduce an additional term in (\ref{eq:totalloss}), based on our optimality analysis. The learning algorithm that we employ for minimizing the loss function is RMSprop \cite{DBLP:journals/corr/abs-1212-5701}, a state-of-the-art stochastic gradient descent algorithm that adapts the learning rate based on a moving window of gradient updates, and is therefore robust to large gradient and noise values.

\section{NDE complexity} \label{sec:complexity}
Ιn this section, we review the complexity of NDE as a design method for LDPC codes. Note that our introduction of NDE does not affect known results about the attractive complexity of this family of codes, related to encoding and decoding \cite{Richardson:2008:MCT:1795974}, as we preserve the sparsity of the code. 

Although the RNN can be unfolded in time to represent the multiple decoding iterations, it only contains two sets of distinct parameters, i.e. the coefficients of $\rho(x)$ mapped to the weights of the first layers and the coefficients of $\lambda(x)$ mapped to the weights of the second layer. This facilitates optimization and suggests that the memory requirements of NDE are negligible. In addition, the convergence rate of RNNs to the fixed point indicated by the labels is geometric and depends on the size of the network, under the assumption that the fixed point is stable \cite{NIPS1988_181}. This suggests that well-performing designs can be found quickly. One pass of back-propagation on the RNN requires time that scales only linearly with the number of (unfolded) layers. We should also note that, in contrast to feedforward networks, this pass cannot be parallelized, due to the time dependency between consecutive layers. The time complexity of previous solutions that employed differential evolution for the design of degree distributions \cite{ 910578} was not investigated, as the focus was that of finding the most well-performing, yet still sub-optimal degree distributions, within the available time budget. In Section \ref{sec:simulations}, we confirm that the time complexity of NDE is superior to differential evolution, and that it scales linearly with the network size.  


\section{Notes on optimality}\label{sec:optimality}
In this following, we describe the optimal solution of code design, analyze the ability of NDE to approach it, and further improve our definition of NDE in order to pursue capacity approaching designs. We characterize a design as capacity-achieving when it ensures successful decoding for noise values arbitrarily close to the performance bound of irregular LDPC codes, $\epsilon_{\text{max}}^{\text{BP}}$ \cite{1077796}. This means that the concept of approaching capacity is not an abstract ability of a design to go sufficiently close to capacity, but follows a closed form expression that defines the optimal solution in relation to code-rate $R$ and the average degree of $\rho(x)$, $\bar{\rho}$.

Based on our definition of the loss function in Section \ref{sec:lossfunction}, we know that, if the loss has been minimized, then the weights correspond to valid degree distributions and the probability of error is very close to zero. This means that DE has a fixed point at 0 for the optimized LDPC code design. However, the probability of failure after decoding can still be high, unless that fixed point is stable. It is known that the dynamic behavior of DE is characterized by the following stability condition, where we have replaced $\epsilon$ in (\ref{eq:DE}) with $x_0$ for the sake of clarity.%

\begin{theorem}(Stability condition for BEC \cite{910578})\label{theo:stabBEC} Let us consider the DE update equation
$$x_t(x_0) = x_0 \lambda(1-\rho(1-x_{t-1}))$$.

Necessity: If $\lambda^{\prime}(0)\rho^{\prime}(1) > 1/x_0$, then there exists a constant $\xi=\xi(\lambda, \rho, x_0)$ such that for all $t \in \mathbb{N}$, $x_t(x_0)>\xi(\lambda, \rho, x_0)$ .

Sufficiency: If $\lambda^{\prime}(0)\rho^{\prime}(1) < 1/x_0$, then there exists a constant $\xi=\xi(\lambda, \rho, x_0)$ such that  if at some $t$ $x_t(x_0) < \xi$ we know that $x_t(x_0)$ converges to 0.
\end{theorem}

Under this stability condition and the observation that the output of NDE is sufficiently low, we know that the design can have zero probability of decoding failure when operating on channels whose erasure values have been used as training data. We take advantage of this condition and we modify the loss function, defined in (\ref{eq:totalloss}), by introducing an additional error term for penalizing weight values that could lead to leaving the stability region of DE:%
\begin{align} \label{eq:updateloss}
    L_{\text{stab}}({y},\hat{{y}}) &= L({y},{\hat{y}}) + c_{\text{stab}} (\lambda^{\prime}(0)\rho^{\prime}(1) - 1/ \epsilon)
\end{align}
where we leverage Theorem \ref{theo:stabBEC} to penalize weight values leading NDE outside the stability region and employ the constant $c_{\text{stab}}$ to control the effect of the new loss term.

In practice, the presence of local minima is not a deal breaker, as long as they are not bad, i.e. the design that they correspond to is sufficiently close to the Shannon capacity. Pursuing capacity-achieving designs faces two major restrictions: \begin{enumerate*}[label=(\roman*)] \item increasing the maximum degrees of the employed degree distributions can help improve performance, but also means that the codeword length needs to be increased. If this does not happen, performance deviates from the asymptotic behavior predicted by DE and the code performs sub-optimally for finite codeword lengths; \item the closer to capacity the channel operates, the more decoding iterations are required. Large codeword lengths are inappropriate for realistic decoder implementations, as they introduce delays and increase complexity \end{enumerate*}. Thus, as we optimize a degree distribution pair $(\lambda(x),\rho(x))$ for channel noise close to capacity and finite decoding iterations, optimization encounters a large number of local minima, an observation also made in \cite{910578}. Local minima are known to attract gradient-based optimizers, which has led to the development of gradient-based variants that employ additional information, such as momentum and second-derivatives, to escape or avoid attraction to bad local optima \cite{DBLP:journals/corr/Ruder16}. However, to the best of our knowledge, there exist no theoretical guarantees for avoiding them and the classical approach is that of performing numerous random initializations of the weight values, which can lead to reaching different local optima of the loss function surface. In the studied problem, we know that, as the weights approach a capacity-achieving design, the loss function surface is characterized by numerous local minima, each one leading to a design that operates very close to capacity. As we commented in Section \ref{sec:intro}, these local optima are not ``bad'', in the sense that they lead to design with performance very close to the optimal one for practical implementations (finite codeword lengths and decoding iterations).

In our implementation, we focus on medium codeword lengths, in the range of $(10^2, 10^5)$ and restrict the maximum degrees of the degree distributions $\lambda(x)$ and $\rho(x)$  to a number significantly lower than the length of the codeword. In addition, we assume finite decoding iterations during the design process, with their number corresponding to the finite number of layers of the unfolded RNN.

We conclude our optimality analysis with a lemma on the sufficiency of simple neuronic units in our formulation of NDE. After the observation in \cite{279181} that RNNs made of simple cells cannot learn long-term dependencies, the machine learning community widely replaced simple cells with the more advanced long short-term memory (LSTM) models. One could, therefore, wonder whether the performance of NDE could be improved if we replace the simple neuronic models used in NDE with LSTM ones. This is not the case here, as DE is a recurrent equation that only exhibits dependencies one step in the past. Therefore, training is free from the common problems of exploding and vanishing gradients caused by long-term dependencies. This observation, which we summarize in Lemma \ref{lemma:simple}, significantly contributes to avoiding the higher complexity of using more advanced models, which can lead to longer training times and difficulties in tuning.

\begin{lemma}(Sufficiency of simple RNN cells)\label{lemma:simple}
The employment of simple neuronic units of the form $y=w \cdot x$, where $y$ denotes the output of a neuron, $x$ its input and $w$ the weight of the input edge, is optimal for NDE.
\end{lemma}

\section{Simulations} \label{sec:simulations}
In this section, we evaluate NDE from various perspectives, primarily aiming at establishing the ability of NDE to find designs that approach capacity and surpass state-of-the-art LDPC code designs for a variety of code configurations in asymptotic settings. All simulations related to the design regard BEC. In addition, we present BER plots for finite codeword lengths over different channel types. Tables \ref{table:setupLearn} and \ref{table:setupCode} summarize the values of parameters used in our simulations (unless different values are explicitly provided in the text) from the perspectives of the RNN and the LDPC code. We have used tensorflow \cite{tensorflow2015-whitepaper} for training NDE were performed using, while some LDPC functionalities were implemented using pylpdc. \footnote{ \url{https://pypi.org/project/pyldpc/0.2/}}

\begin{table}[t]
\begin{minipage}[t]{0.45\textwidth}
\centering
\caption{Simulation setup: RNN hyperparameters}
\scalebox{0.7}{
\begin{tabular}{|c|c|}
\hline
Simulation parameter & value\\
\hline
$E$, training epochs & 200 \\ 
$D_{\text{train}}$, number of training instances & 11220  \\
$D_{\text{test}}$, number of testing instances & 2783  \\
$\alpha$, learning rate & 0.0001 \\
$N$, number of layers (of the unfolded RNN) & 10 \\
$o_{\text{low}}$, output threshold & 0.0009 \\
$c_{\text{MSE}}, c_{\lambda}, c_{\rho}, c_{\bar{\lambda}}, c_{\text{stab}}$, loss function constants & 100, 6, 1, 100  \\
 \hline
\end{tabular}}
\label{table:setupLearn}
\end{minipage}  \quad\quad\quad
\begin{minipage}[t]{0.45\textwidth}
\centering
\caption{Simulation setup: LDPC code parameters}
\scalebox{0.7}{
\begin{tabular}{|c|c|}
\hline
Simulation parameter & value\\
\hline
$k$, message size & 1024 \\
$R$, code-rate & 0.5 \\
$\lambda_{\text{max}}$, maximum variable node degree & 15\\
$\rho{\text{max}}$, maximum check node degree & 12\\
$d$, density &  0.01 \\
\hline
\end{tabular}}
\label{table:setupCode}
\end{minipage}
\end{table}

\subsection{Capacity approaching designs}
We have simulated a variety of well-performing code designs from the related literature, including both regular \cite{Richardson:2008:MCT:1795974} and irregular \cite{Luby:1998:ALD:276698.276756} LDPC codes and consolidated the results in Table \ref{table:results}, where we also include our best-performing designs, i.e. NDE$_1$. The majority of the codes presented in Table \ref{table:results} exhibit a rate of approximately 0.5, as it is commonly studied in the literature. In addition, we also evaluate NDE on the design of codes with $R \in \{ 0.33, 0.8\}$. Table \ref{table:results} presents the results of these simulations, where we can see that NDE has approached capacity for a variety of configurations. For each design, we present the characteristics of the degree distributions and a number of metrics related to the performance of the code, where we define the gap $\delta$ as the difference $\epsilon^{\text{Sh}} - \epsilon^{\text{BP}}$ and employ this metric to evaluate the code design in an asymptotic setting.

\begin{table}
\centering
   \begin{minipage}[t]{0.75\textwidth}
\caption{Evaluation of different design methods: regular LDPC codes produced in \cite{Richardson:2008:MCT:1795974} (Table 3.59), irregular LDPC codes produced in \cite{910578} (Table 1) and irregular LDPC codes produced using NDE.}
\label{table:results}
\centering
\scalebox{0.95}{
\begin{tabular}{|l|l|l|l|l|l|l|}
\hline
 Method & $\bar{\lambda}$  & $\bar{\rho}$  & $R$  & $\epsilon^{\text{Sh}}$ & $\epsilon^{\text{BP}}$ & $\delta$\\
 \hline
 $\text{R}_{1}$  \cite{Richardson:2008:MCT:1795974} & 3  &  6 & 0.5  & 0.5 &  0.4297 & 0.0703\\
 \hline
$\text{R}_2$  \cite{Richardson:2008:MCT:1795974} &  4 & 8 & 0.5 & 0.5 &  0.3837  & 0.116\\
\hline
$\text{IR}_{4}$ \cite{910578} & 2.8619 & 4.7156 & 0.3930 & 0.6069 & 0.5603 & 0.0466\\
\hline
$\text{IR}_{9}$ \cite{910578} & 3.5492 & 7.0984 & 0.5000 &  0.4999 & 0.4711 & $\bm{0.0288}$\\
\hline
$\text{IR}_{12}$ \cite{910578} &  3.8627 & 7.7280 & 0.5001 & 0.4998 & 0.4684 & 0.0313 \\
\hline
$\text{NDE}_{1}$ &  3.6043 & 7.1470 & 0.4956 & 0.4956 & 0.4916 & $\bm{0.01268}$ \\
\hline
$\text{NDE}_{2}$ & 2.6196 & 3.8604 & 0.3214 & 0.6786 & 0.6343 & 0.0442\\
\hline
$\text{NDE}_{3}$ & 2.5780 & 12.7832 & 0.7982 & 0.2018 & 0.1705 & 0.031 \\
\hline
\end{tabular}}
\end{minipage}
\end{table}

\begin{figure}
\begin{minipage}[t]{0.45\textwidth}
\includegraphics[scale=0.22]{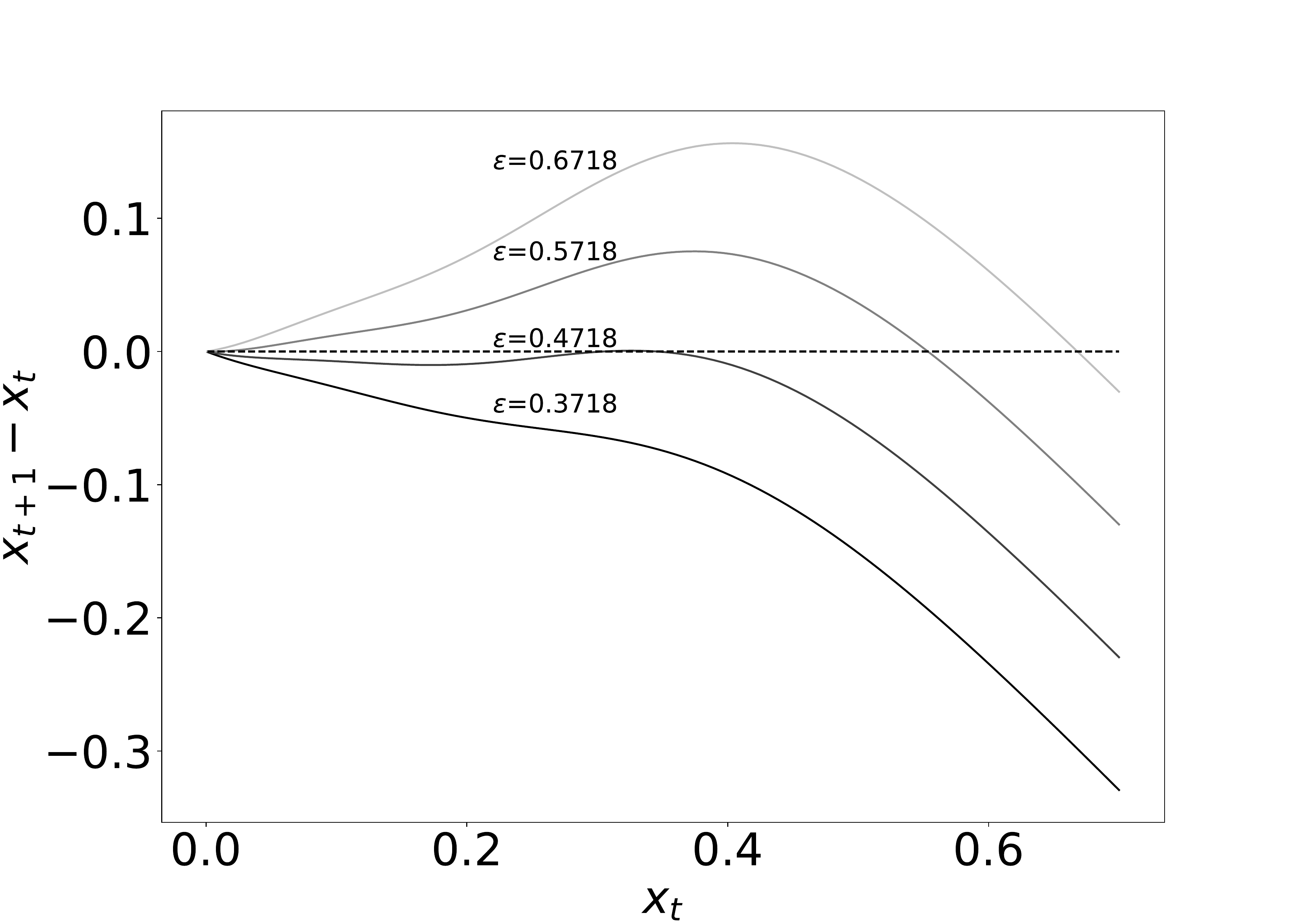}
\caption{Graphical determination of threshold for IR$_{9}$ (see Table \ref{table:results}).}
\label{fig:ebp_baseline}
\end{minipage} \quad\quad\quad
\begin{minipage}[t]{0.45\textwidth}
\includegraphics[scale=0.22]{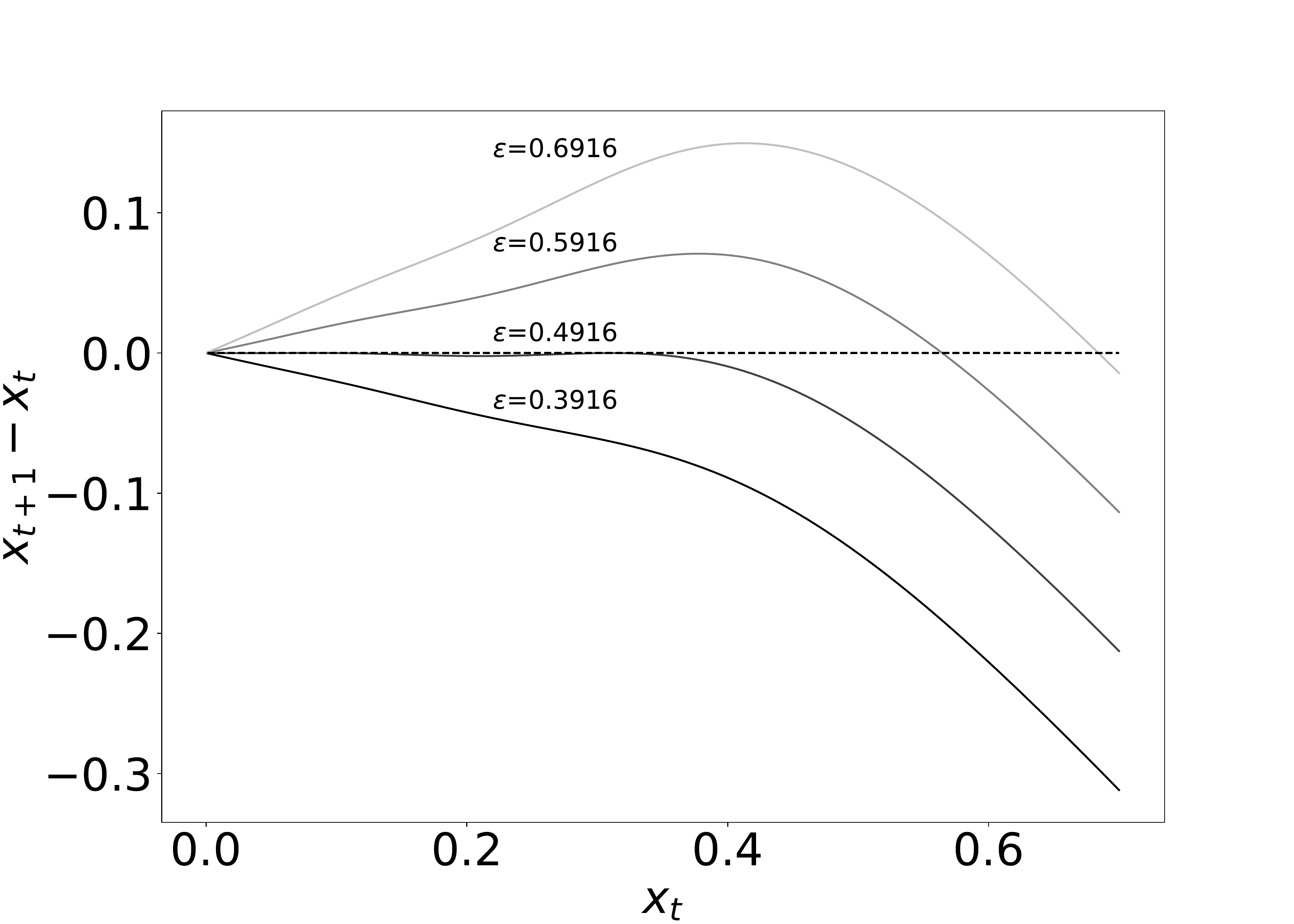}
\caption{Graphical determination of threshold for NDE$_{1}$ (see Table \ref{table:results}).}
\label{fig:ebp_ours}
\end{minipage}
\end{figure}

Figs. \ref{fig:ebp_baseline} and \ref{fig:ebp_ours} depict the graphical determination of the threshold of the code design for varying erasure values $\epsilon$ for the best-performing baseline method, $\text{IR}_{9}$, and our solution $\text{NDE}_1$, respectively. In this type of plot, $\epsilon^{\text{BP}}$ is calculated as the maximum value of $\epsilon$ for which the graph of $x_{t+1}-x_t, \ x_t \in [0,1]$ does not exceed 0. When this graph tangentially touches 0 for a particular $x_t$ (critical point), then the decoder halts, as it cannot decrease the probability of decoding failure further. As was observed in \cite{Richardson:2008:MCT:1795974}, the more optimized the design is, the more critical points the graph will exhibit. This is due to the fact that, when a code is capacity-approaching, all points $x \in [0, \epsilon^{\text{BP}}]$ are critical points. As can be seen in Fig. \ref{fig:ebp_ours}, in addition to exhibiting higher $\epsilon^{\text{BP}}$, our design exhibits a significantly larger number of critical points, so that seemingly all points in $x \in [0,0.49]$ are fixed points. 

\subsection{Insights into optimization}
\begin{figure}
\begin{minipage}[t]{0.45\textwidth}
\includegraphics[scale=0.22]{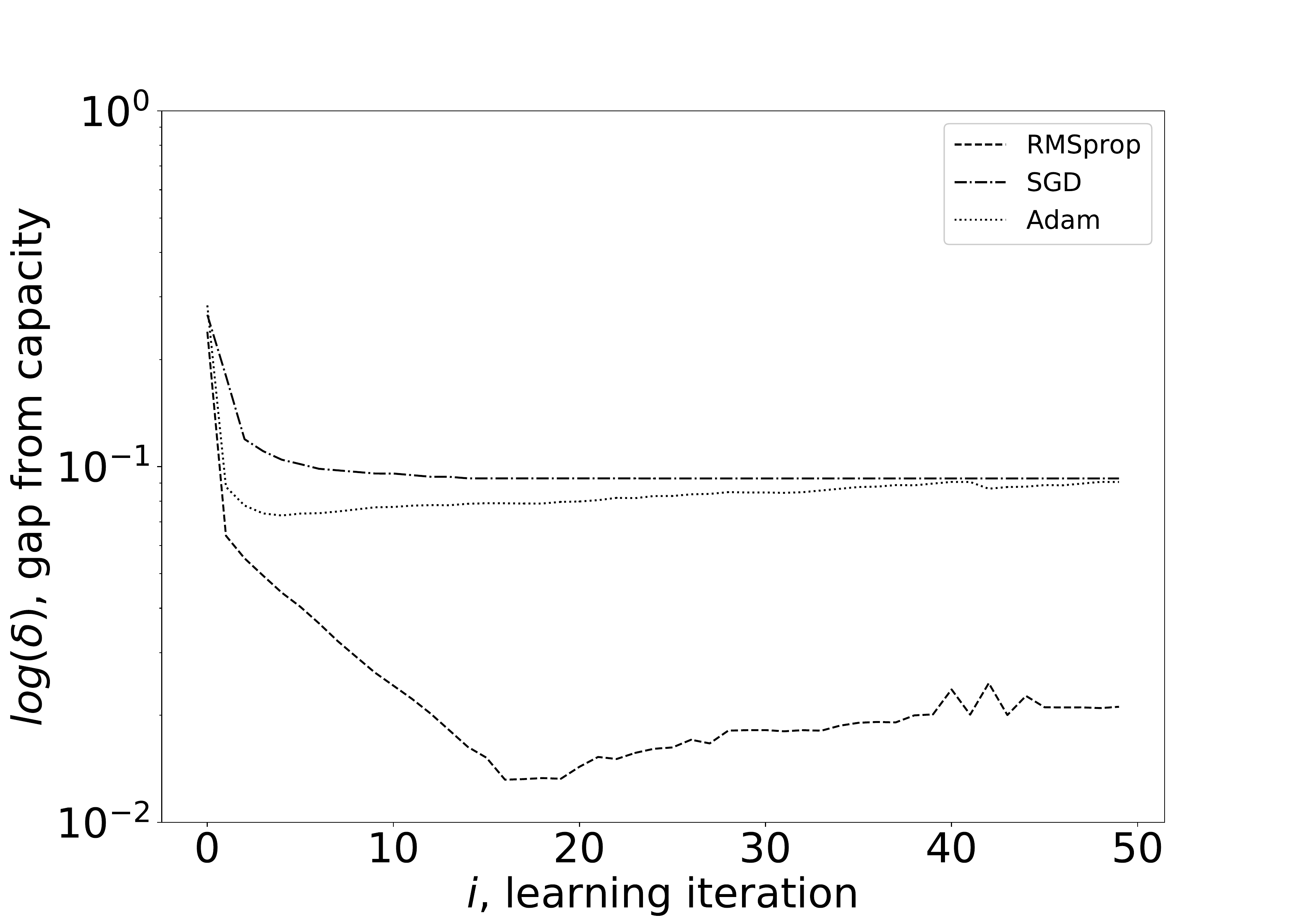}
\caption{Evolution of $\epsilon^{\text{BP}}$ with training time for different optimizers.}
\label{fig:optimizers}
\end{minipage} \quad\quad\quad
\begin{minipage}[t]{0.45\textwidth}
\includegraphics[scale=0.22]{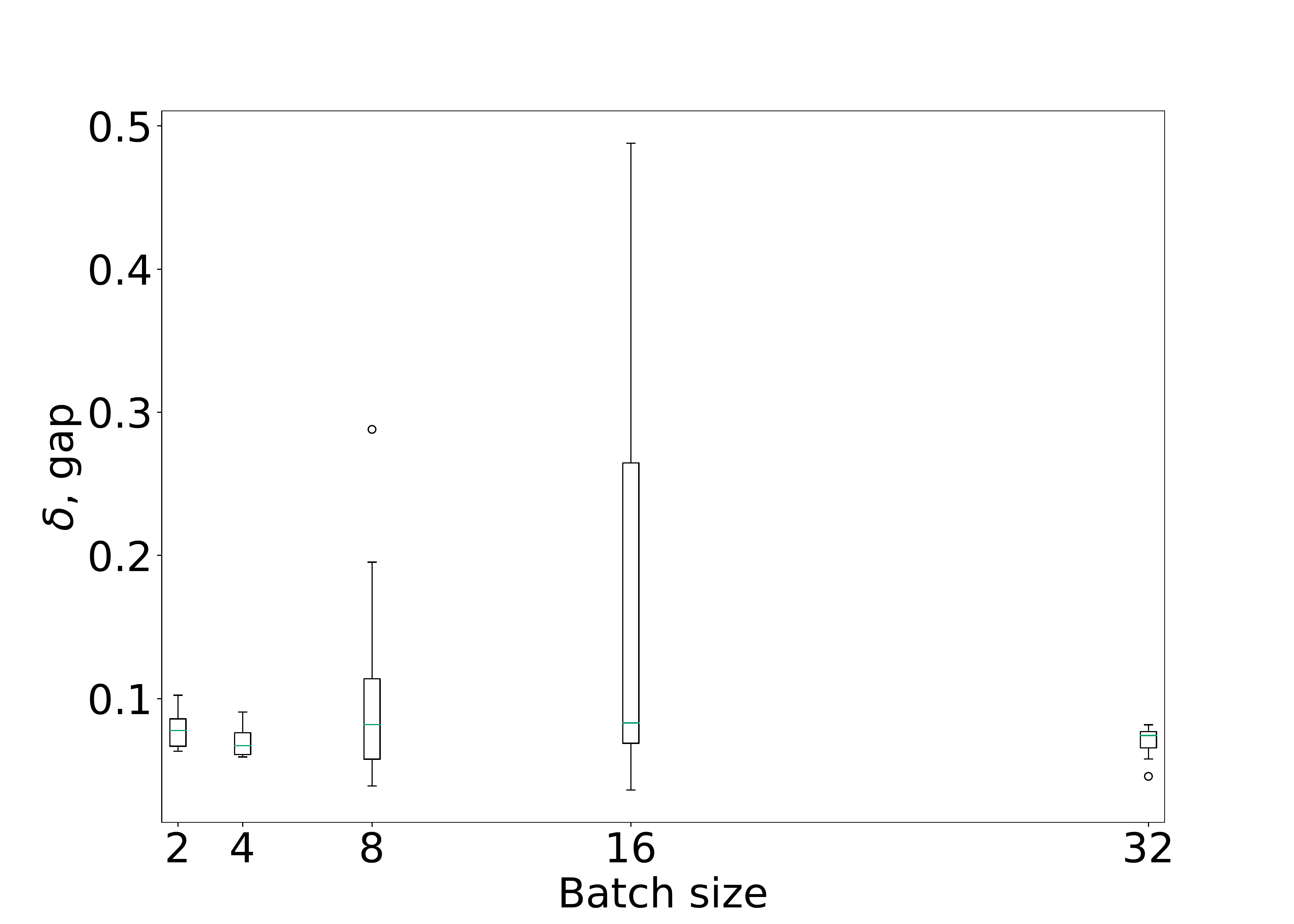}
\caption{Achieved gap $\delta$ for varying batch sizes and random initialization of weights.}
\label{fig:batches}
\end{minipage}
\end{figure}

In this section, we shed light on the training procedure of NDE in order to get a better understanding of how the different hyperparameters of the learning algorithm and the RNN architecture affect the quality of the learned degree distributions. In Fig. \ref{fig:optimizers}, we present the evolution of the gap $\delta$ with training iterations, to examine the impact of the solver on the derived solution and convergence. Note that, we do not plot the values of the training loss, as it consists of many terms (see Eq. (\ref{eq:totalloss})) and exhibits variations that depend on the choice of the output threshold $o_{\text{low}}$. We instead depict the gap $\delta$, as it is a clear indicator of the quality of the design. However, we would like to emphasize that the calculation of $\epsilon^{\text{BP}}$ takes place only during the testing trials and not during the training ones, as this information is not part of the loss function. This helps us to avoid increasing the time complexity of training. If we focus on the best-performing model in  Fig. \ref{fig:optimizers}, which employs RMSprop, we observe that $\delta$ is very high at the beginning of the training process, as the weights of NDE do not correspond to valid degree distributions (see conditions in (\ref{eq:constraints})), but decreases significantly after 10 training iterations. Next, optimization focuses on improving the asymptotic performance of the design and, as we can observe in Fig. \ref{fig:optimizers}, the training procedure initially quickly approaches a small value, and then convergence to the final design is slow. After reaching the lowest value at 190 iterations, the distance starts increasing again. This can be attributed to the fact that the derivative values after 190 iterations are not high enough, due to the optimization algorithm having reached a local optimum. Finding a better solution is then hard, so the optimization algorithm starts moving towards worse solutions. As the learning rate has been decreased to low levels, the procedure cannot find the good solution again. Based on this observation, if we observe that testing performance deteriorates, we terminate training early.  

We also present results that show the effect of different hyperparameterizations, in particular the choice of the optimizer and batch size. This analysis has the twofold objective of revealing how tuning was performed and also drawing insights into the nature of the code design problem. We begin by examining the performance of different optimizers in Fig. \ref{fig:optimizers}. Note that our objective is not that of searching for the best-performing optimizer among all optimizers in the literature \cite{DBLP:journals/corr/Ruder16}, as this would require exhaustive tuning of each one. Instead we focus on investigating whether classical gradient descent is capable of reaching good solutions or gets stuck at local optima. We experimented with RMSprop and Adam, two state-of-the-art gradient-based algorithms designed to effectively avoid local optima. In Fig. \ref{fig:optimizers}, we present the performance of these three optimizers. For the simple gradient-descent (SGD) optimizer, we have found the best value for the learning rate $\alpha$ using grid-search in the interval $(0.00001, 0.001)$. The same weight initialization and RNN architecture was used for all three optimizers. We kept the hyperparameters of Adam to their default values, as these were set by the authors in \cite{kingma2014adam}. It is possible that tuning these parameters would give better results, but for these simulations we only tuned the learning rate and concluded that a value of 0.0001 performs optimally. As RMSprop exhibits good performance without requiring extensive tuning, we employ it in the remaining simulations. 

In general, the batch size $b$ has a significant effect on the performance of an optimizer, with smaller sizes favoring fast convergence and helping avoiding local minima, while larger sizes making optimization more robust to noise and efficient in parallelized implementations. When using batches in an online learning setting (instead of feeding the whole dataset at each learning iteration), the number of training data does not matter, but convergence depends only on the number of updates (learning iterations per episode) and the richness of the training distribution \cite{DBLP:journals/corr/abs-1206-5533}. As we are creating artificial data, the task can be seen as online learning. During tuning we aim at finding a well-performing combination of learning rate and batch size, with larger batch sizes resulting in fewer learning iterations per epoch. Fig. \ref{fig:batches} shows the performance achieved using a random initialization of the RNN weights. We observe that the different batch sizes exhibit similar mean performance, with a size of 4 achieving the smallest gap, but variance differs significantly. In particular, the variances for a batch size of 8 and 16 are both very high. When $b$ is set to 8, the optimizer in most cases satisfies the structural requirements but cannot optimize the threshold $\epsilon^{\text{BP}}$, while for $b=16$ the structural requirements are often not satisfied due to selecting weights that do not correspond to valid degree distribution. These numerical results depend highly on the previously selected value of the learning rate.

\subsection{Dynamic analysis of degree distributions}
\begin{figure}
\begin{minipage}{0.45\textwidth}
    \centering
\includegraphics[scale=0.22]{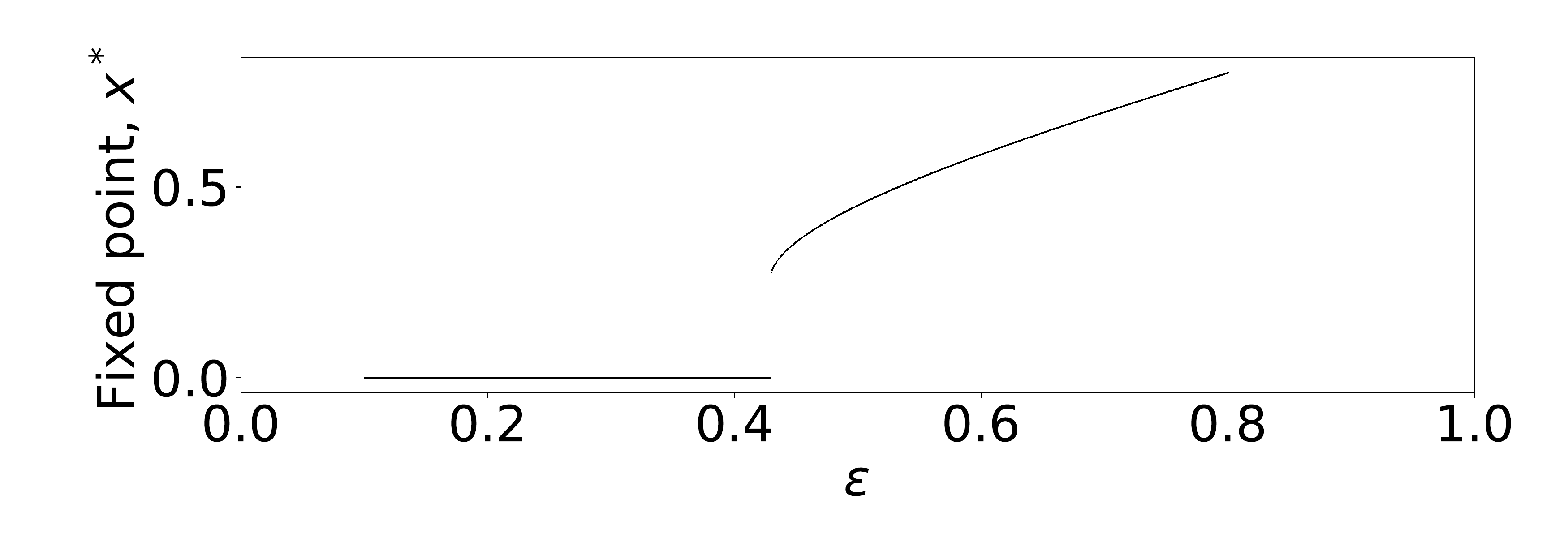}
    \vspace{0.05cm}
    \includegraphics[scale=0.22]{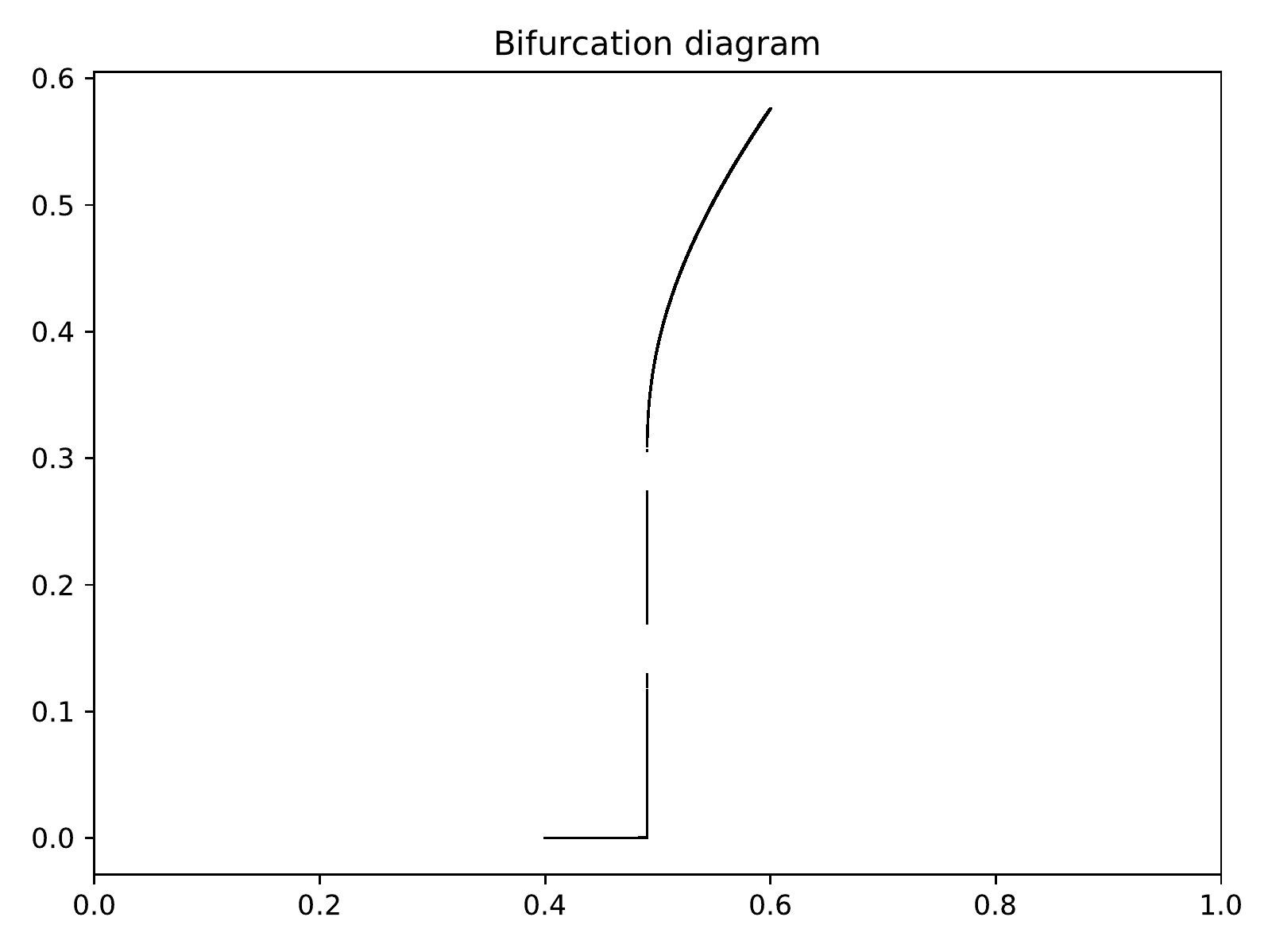}
    \caption{Bifurcation diagrams of our best-performing solution and R$_2$.}
    \label{fig:bifurcNDE}
\end{minipage} \quad \quad \quad
\begin{minipage}{0.45\textwidth}
    \centering
    \includegraphics[scale=0.14]{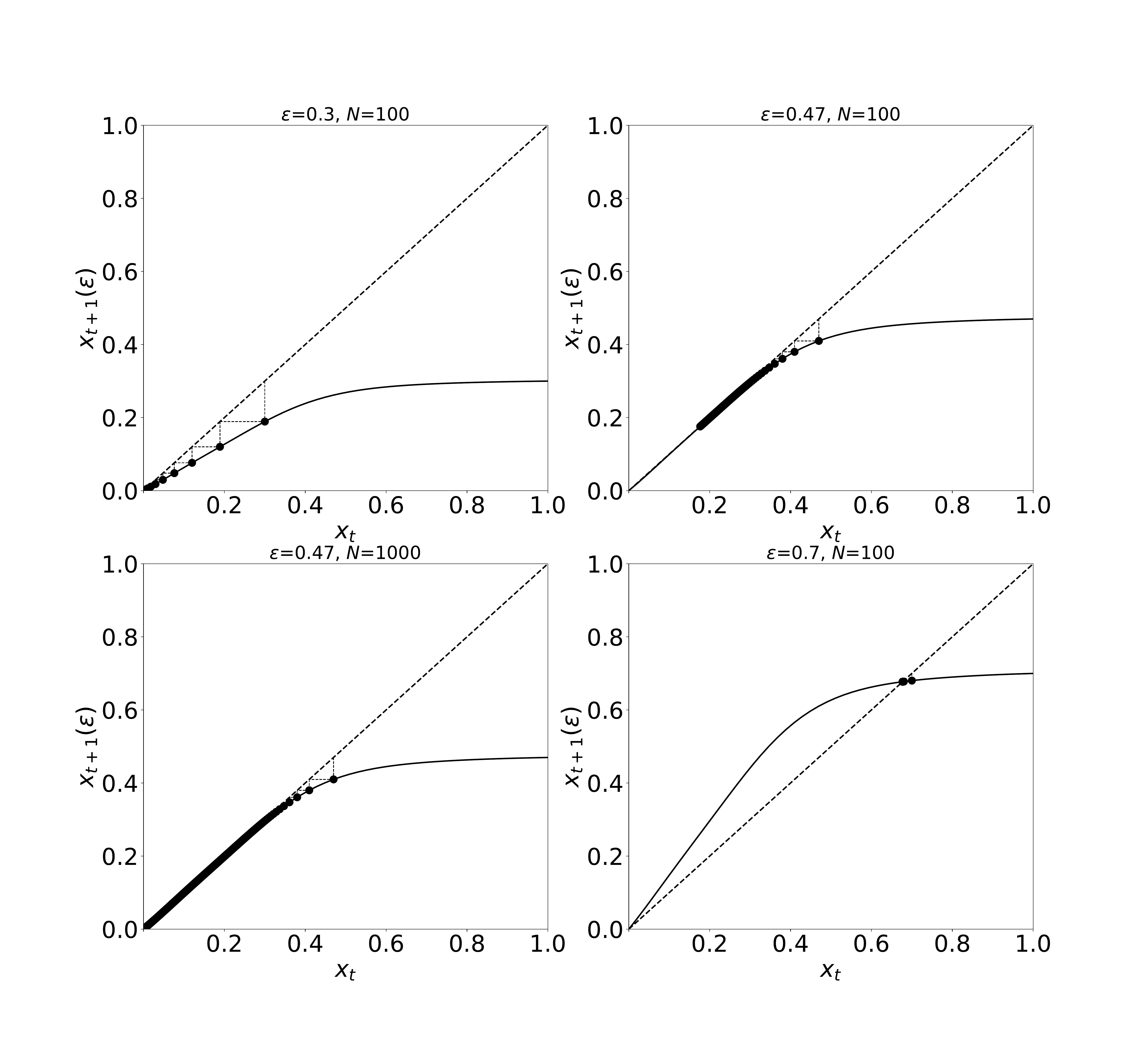}
    \caption{Dynamic behavior of NDE for different erasure values.}
    \label{fig:dynamicNDE}
\end{minipage}
\end{figure}

In this section, we employ bifurcation diagrams to gain a better understanding of the dynamic behavior of our NDE-based LDPC code designs. In particular, Fig. \ref{fig:bifurcNDE} contains the bifurcation diagrams for our best-performing design $\text{NDE}_1$ and the baseline $\text{R}_2$. We observe that, for NDE$_1$, almost all points close to capacity can be fixed points when the initial point $\epsilon$ equals the Shannon capacity, whereas $\text{R}_2$, which was chosen as the baseline method in Table \ref{table:results} with the worst performance, does not exhibit this behavior. Furthermore, in Fig. \ref{fig:dynamicNDE}, we present the convergence of DE to a fixed point for channels of increasing noise values. We observe that, when the experienced noise is low ($\epsilon=0.3$), only a few iterations are required, while when we get closer to capacity ($\epsilon=0.47$), we need to increase the number of decoding iterations from 100 to 1000 to ensure convergence to 0. Finally, for erasure values above the Shannon capacity ($\epsilon=0.7$), DE converges very close to the initial point, which suggests that decoding has failed. This behavior has already been observed in Fig. \ref{fig:bifurcNDE}, where all points above $\epsilon^{\text{Sh}}$ are fixed points.

\subsection{Increasing the RNN size}
The size of the RNN is determined by the number of neurons in each layer, as the number of layers is constrained to 2 by the definition of the architecture and the number of neurons in the first (second) layer agrees with the maximum degree of the degree distribution $\lambda(x)$ ($\rho(x)$). Thus, when changing the size of the RNN we adjust the average degrees $\bar{\lambda}$ and $\bar{\rho}$ accordingly.

\begin{figure}
\begin{minipage}[t]{0.45\textwidth}
\centering
   \scalebox{0.22}{
    \includegraphics{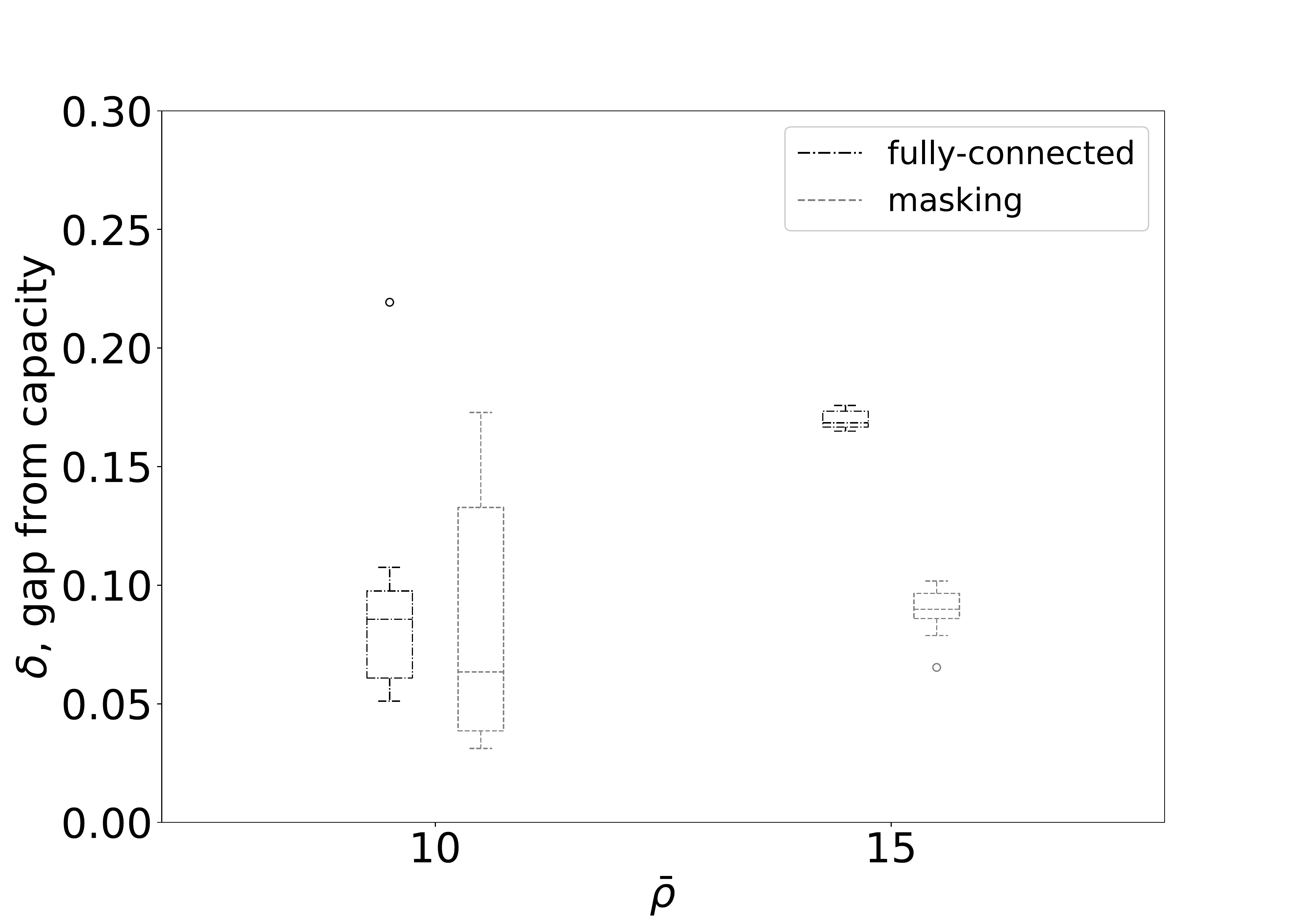}}
    \caption{Comparison of the achieved gap $\delta$ for a fully-connected RNN and an RNN employing masking for varying $\bar{\rho}$.}
    \label{fig:averages}  
\end{minipage} \quad\quad\quad
\begin{minipage}[t]{0.45\textwidth}
\centering
   \scalebox{0.22}{
    \includegraphics{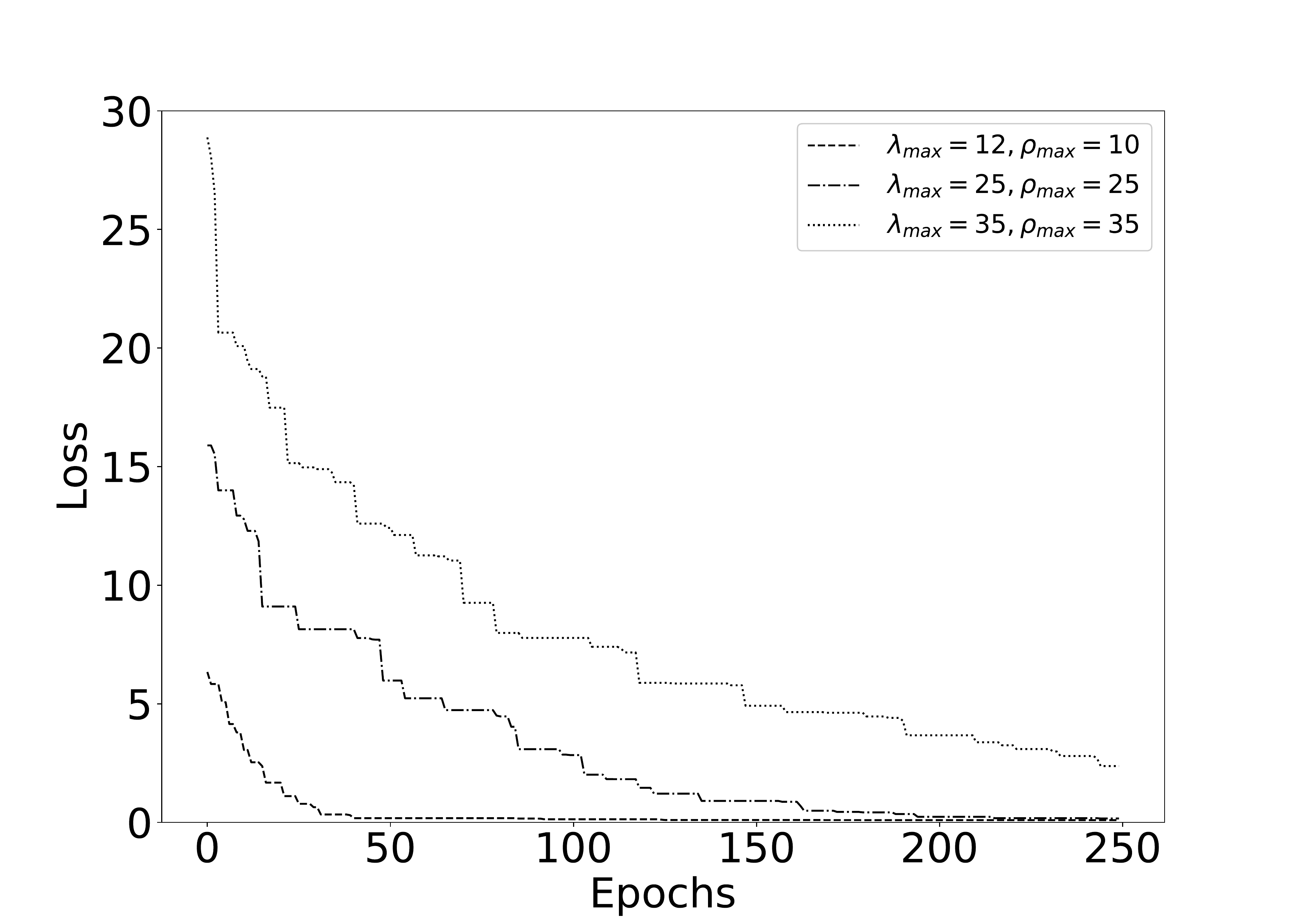}}
    \caption{The evolution of the loss using Differential Evolution for a varying number of degree coefficients.}
    \label{fig:diffE}  
\end{minipage} 
\end{figure}

In Fig. \ref{fig:averages}, we present the achieved $\epsilon^{\text{BP}}$ of NDE for varying average and maximum degrees. We, also, examine different combinations of maximum degrees and percentages of zero coefficients and perform random initializations of the RNN weights. When increasing $\bar{\rho}$, we made two observations: \begin{enumerate*}[label=(\roman*)]
    \item depending on the previous values, we will also need to increase the maximum degree of $\rho(x)$, and perhaps also $\lambda(x)$. We usually keep these values as small as possible, to reduce the dimensionality of the problem and decrease computational time. As increasing the number of coefficients makes the search of the optimal coefficient values harder, we introduced weight masking in the training process. This technique was also followed in \cite{910578}, as it was observed that very good distribution pairs exist with only a few non-zero terms. At the beginning of training, we randomly set to 0 a predefined percentage of the weights and also nullify the corresponding gradients during training;  
    \item randomness decreases. This can be attributed to the fact that the number of optimization parameters increases and, therefore, the probability of starting the search from a ``bad'' random initialization is lower.
\end{enumerate*} In Fig. \ref{fig:averages}, we experiment with two types of RNN:
\begin{enumerate*}[label=(\roman*)]
    \item a fully-connected RNN where the number of trainable weights is equal to $\lambda_{\text{max}} + \rho_{\text{max}}$ (30 for $\bar{\rho}=10$ and 40 for $\bar{\rho}=15$);
    \item an RNN that employs masking. In particular, $30\%$ of the weights are randomly selected for masking and the total number of weights is 40 and 50 for $\bar{\rho}=10$ and $\bar{\rho}=15$, respectively
\end{enumerate*}. We observe that the performance achieved by a fully-connected network is in both cases worse, with performance significantly deteriorating as $\bar{\rho}$ increases. This can be attributed to the large number of weights that make it hard for the optimizer to nullify all loss terms. When employing masking, performance slightly deteriorates and variance decreases when $\bar{\rho}$ increases. This slight deterioration can be attributed to insufficient tuning of the optimizer.

\subsection{Efficiency}\label{sec:efficiency}
In this section, we evaluate the efficiency of NDE in terms of time complexity and compare it with that of differential evolution, the optimization method used by the baseline designs presented in Table \ref{table:results}. Our implementation follows the description in \cite{910578}, but we did not aim at replicating the presented results. Instead,we focus on measuring how the time complexity  and convergence rate of differential evolution depends on the problem size, i.e. the number of coefficients. We have defined the loss function minimized by differential evolution in a similar spirit to our work, i.e. we optimize the decoding threshold $\epsilon^{\text{BP}}$ and ensure that the found coefficients correspond to degree distributions. In particular, the loss function is $ L_{\text{DiffE}} = |\epsilon^{\text{Sh}} - \epsilon^{\text{BP}}| + |R - (1-\bar{\lambda} - \bar{\rho})| + |1-\sum_{i=0}^{\lambda_{\text{max}}}{\lambda_i}| + |1-\sum_{i=0}^{\rho_{\text{max}}}{\rho_i}|$. 

\begin{figure}
   \centering
    \begin{minipage}[t]{0.45\textwidth}
    \scalebox{0.22}{
    \includegraphics{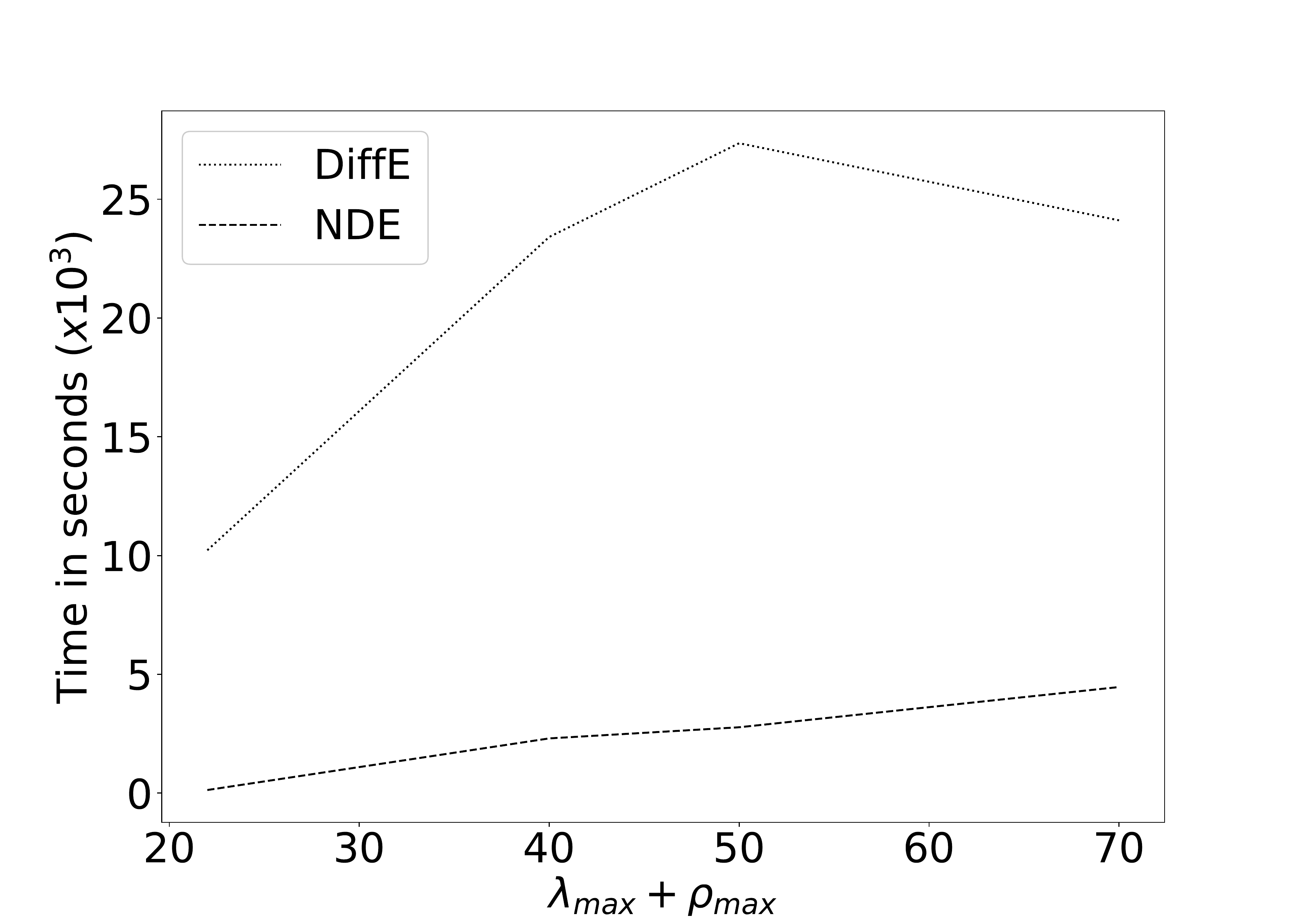}}
    \caption{Comparison of time complexity of Differential Evolution and NDE.}
    \label{fig:times}
\end{minipage} \quad\quad\quad
\begin{minipage}[t]{0.45\textwidth}
             \centering
        \scalebox{0.22}{
    \includegraphics{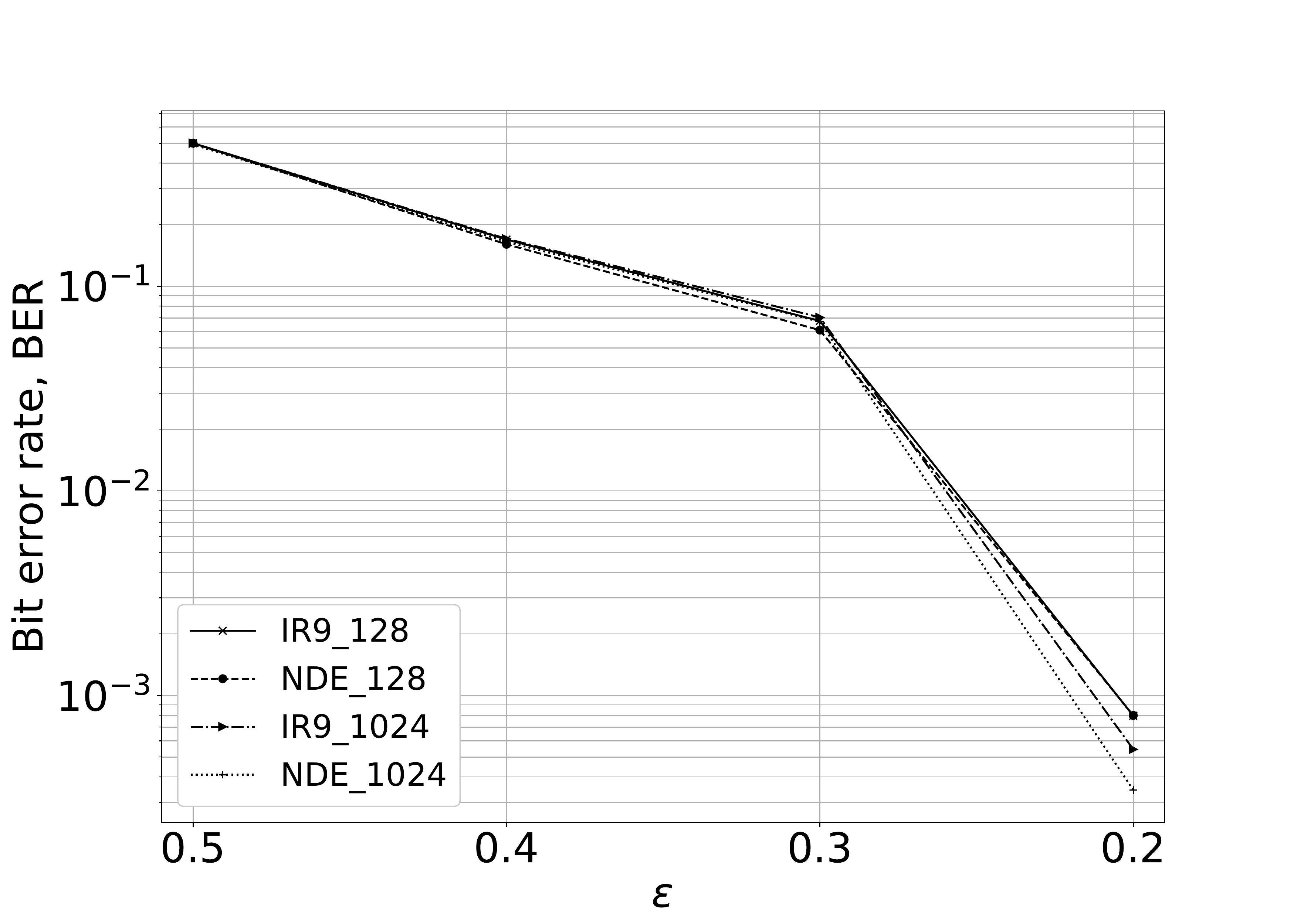}}
    \caption{BER achieved for transmission over the binary erasure channel for our best-performing design ($\text{NDE}_1$) and the baseline ($\text{IR}_9$) for messages with length $k=128$ and $k=1024$ and code-rate $R=0.5$.}
    \label{fig:bec_compare} 
\end{minipage} 
\end{figure}

In Fig. \ref{fig:diffE}, we present the evolution of the loss with the number of epochs for differential evolution, where we allowed a time budget of 250 epochs. We also experiment with different maximum degrees to examine the effect of the problem size on the quality and speed of differential evolution. As regards speed, we can see that, when the number of coefficients is small, differential evolution converges at around 50 epochs. Doubling the number of coefficients makes the problem harder, thus around 150 iterations are required to reach convergence. Finally, increasing the number of coefficients to 70 leads to designs that give non-zero loss. Note that, we do not compare these results with the ones presented in Fig. \ref{fig:optimizers} related to the convergence speed of our method. This is because the definition and complexity of epochs in machine learning and evolutionary algorithms differs significantly. Instead, we compare the two approaches in Fig. \ref{fig:times} in terms of absolute running time and observe that the running times of both NDE and differential evolution increase in a linear way, but differential evolution requires around 5 times the time required by NDE for all problem sizes. We should note that, the reason why the time complexity of differential evolution stopped increasing for problems with more than 50 coefficients, is that it exceeded its time budget. As we can see in Fig. \ref{fig:diffE}, this results in failing to reach zero loss for a problem of size 70.

\subsection{Bit-error-rate evaluation}
In this section, we evaluate the performance of our design for finite codeword lengths in order to examine whether the superiority of NDE, already established in an asymptotic analysis, still remains. We construct our codes randomly, by sampling the produced degree distributions. As was described in \cite{Richardson:2008:MCT:1795974}, we calculate the number of edges as $ e = \bar{\lambda}=\bar{\rho} $ and assign a label to each edge from the set $ S_1 = \{1, 2, \cdots, e \} $. We define $e$ sockets both for variable nodes check nodes, where each variable (check) node has a number of sockets chosen by sampling $\lambda(x) (\rho(x))$. We, then, take a random permutation of $S_1$, which we denote as $S_2= \{ \pi(1), \pi(2), \cdots, \pi(e) \}$, where $\pi(\cdot)$ indicates a random permutation of a label. Finally, we connect each edge from set $S_1$ to the corresponding element in $S_2$ and remove all 4-cycles. 

\begin{figure}
\centering
\begin{minipage}[t]{0.95\textwidth}
        \centering
\scalebox{0.22}{\includegraphics{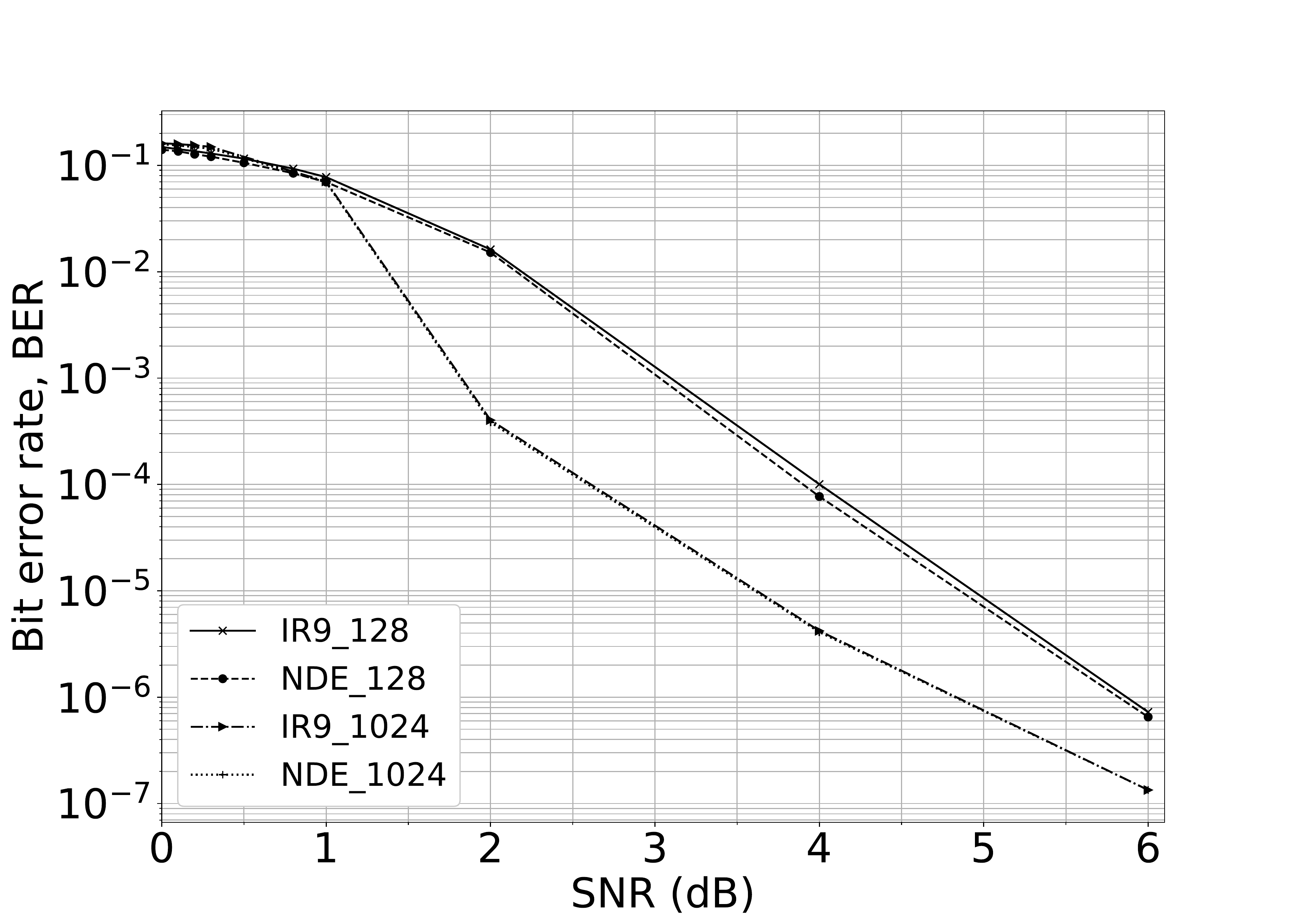}}
    \caption{BER achieved for transmission over the AWGN channel for our best-performing design ($\text{NDE}_1$) and the baseline ($\text{IR}_9$) for messages with length $k=128$ and $k=1024$ and code-rate $R=0.5$.}
    \label{fig:awgn_compare}

\end{minipage}
\end{figure}

Although our NDE formulation concerns the BEC channel, we anticipate, based on empirical observations in \cite{910578}, that our designs will perform well for a variety of channels. We therefore evaluate the performance of $\text{NDE}_1$ in the BEC and the AWGN channel and compare it with the performance achieved by $\text{IR}_9$, the best-performing baseline method. Results are shown in Figs. \ref{fig:bec_compare} and \ref{fig:awgn_compare}. As we anticipated, the performance of the two designs does not differ significantly. In Fig. \ref{fig:bec_compare}, we observe that for erasure values close to 0.5, which corresponds to the Shannon capacity of this code, BER remains close to 0.5. As the channel becomes less erroneous, BER improves for both designs. In Fig. \ref{fig:awgn_compare}, we observe that performance improves drastically for large codewords ($n=2048$) after 1 dB for both designs. In general, we can conclude that our design is well-performing across all channels under consideration and that shorter codewords exhibit worse performance than longer ones, which is aligned to what has been observed in the literature. 


\section{Conclusions and future work} \label{sec:discussion}
In this work, we introduced neural density evolution, which models density evolution in the binary erasure channel as a recurrent neural network, and employed it for the design of irregular LDPC codes. Our simulations indicate that our designs achieve, and in some cases exceed, the performance of state-of-the-art designs in asymptotic settings and are well-performing for a variety of codeword lengths and channels. In addition, we have confirmed that NDE exhibits lower complexity than its counterparts while it avoids bad designs. Moreover, our loss function allows for the design of codes with variable rates, and can be further customized based on the code's structural requirements. We believe that our methodology can bring further benefits for channel coding, when combined with a neural network architecture that models DE for arbitrary channels. This tool could then offer capacity-approaching designs without requiring a model of the channel, but solely based on training data consisting of the original messages and the messages after transmission through a channel. Our approach can be beneficial for other coding schemes, such as cascades of codes and 5G New Radio codes, as it can render the design of these types of irregular LDPC codes efficient and customizable.





\bibliographystyle{IEEEtran}

\end{document}